\documentclass[]{openmoss}

\usepackage{tabularx}
\usepackage{array}
\usepackage{booktabs}
\usepackage{multirow}
\usepackage{graphicx}
\usepackage{amsmath}
\usepackage{amsfonts}
\usepackage{amssymb}
\theoremstyle{plain}
\newtheorem{proposition}{Proposition}
\newtheorem{lemma}{Lemma}
\newtheorem{corollary}{Corollary}
\theoremstyle{remark}
\newtheorem{remark}{Remark}
\theoremstyle{plain}
\usepackage{nicefrac}
\usepackage{algorithm}
\usepackage{algpseudocode}
\makeatletter
\providecommand*{\toclevel@algorithm}{0}
\makeatother
\usepackage{enumitem}
\usepackage{listings}
\usepackage{stfloats}
\usepackage{placeins}
\usepackage[table]{xcolor}
\usepackage{arydshln}
\usepackage{marvosym}
\usepackage{pifont}
\usepackage{tcolorbox}
\tcbuselibrary{breakable}

\newcommand{\improvement}[1]{#1}

\newcommand{\ours}{SEAL}
\newcommand{\stitle}[1]{\paragraph{#1}}
\newcommand{\cone}{\ding{182}}
\newcommand{\ctwo}{\ding{183}}
\colorlet{tablegray}{gray!8}
\colorlet{tableaccent}{orange!10}
\title{SEAL: Synergistic Co-Evolution of Agents and Learning Environments}

\author[1,2]{Yihao Hu}
\author[\Letter,1]{Zhihao Wen}
\author[1,4]{Pan Wang}
\author[3]{Xiujin Liu}
\author[1]{Yitong Lin}
\author[1]{Xin Zhang}
\author[1]{Wei Wu}

\affiliation{%
\parbox{\textwidth}{\centering
    \affiliationmark{\Letter} Corresponding Author (\email{z.wen@antgroup.com}) \\
    \affiliationmark{1} Ant Group \quad
    \affiliationmark{2} Westlake University \\
    \affiliationmark{3} University of Michigan--Ann Arbor \quad
    \affiliationmark{4} University of Science and Technology of China \quad
}}

\abstract{
Large Language Model (LLM) agents increasingly learn through interaction, yet their training process often evolves unevenly: the policy changes with experience while the learning environment continues to provide largely fixed forms of support. As training progresses, this creates a growing mismatch between what the agent currently struggles with and what its learning process emphasizes. We call this phenomenon \emph{Agent--Learning Environment Misalignment} (ALEM). We introduce \ours, a framework for \emph{synergistic co-evolution} of agents and learning environments. \ours\ turns verifier-grounded execution failures into a shared diagnostic signal that coordinates two complementary forms of adaptation: it reshapes the support available in subsequent interactions and adjusts how the policy learns from collected trajectories. In this way, the learning environment follows the agent's evolving failure profile while both sides remain grounded in the same executable feedback. Across three backbones on BFCL V3, \ours\ consistently improves over matched Vanilla RL, reaching 40.25\% versus 30.75\% with Qwen2.5-7B. A matched factorial ablation further yields a +5.00-point interaction estimate for the coupled configuration. The same trained checkpoints also show positive transfer to the evaluated BFCL V4 and $\tau^2$-bench domains. These results suggest that coordinating how an agent learns with the environment it learns from is a promising direction for self-evolving tool-use agents.

}
\checkdata[Homepage]{\url{https://yihaohu0118.github.io/SEAL/}}
\checkdata[Github Repo]{\url{https://github.com/yihaohu0118/SEAL}}

\begin{document}
\maketitle

\section{Introduction}
\label{sec:intro}

Ask an LLM agent for the cost of an economy flight from Rivermist to Stonebrook. It guesses the airport codes, calls \texttt{get\_flight\_cost} with \texttt{travel\_from="RVM"} and \texttt{travel\_to="STB"}, and the environment answers with a generic execution error, \emph{No available route for the given airports}; the agent concludes that no route exists and gives up (Figure~\ref{fig:case_main}). Under standard reinforcement learning (RL), that is the whole lesson: the rollout receives reward zero, and the next rollout is collected under exactly the same conditions. Nothing records whether the agent skipped the airport-code lookup, passed an invalid argument, or failed to recover, so neither the policy update nor the environment can act on the difference. This loop is now a dominant way LLM agents improve. Beyond their capabilities in reasoning, planning, and tool use \citep{yao2022react, schick2023toolformer, shinn2023reflexion, chen2026omnivideo, yu2026dual}, agents are refined through RL, tool-use post-training, agentic data generation, and multimodal data evolution \citep{qian2025toolrl, wei2026agentic, hu2025agentgen, gao2026counterfactual}, and more generally through \emph{self-evolution through interaction}: collecting rollouts, receiving feedback, and improving via reflection, RL, self-generated supervision, or skill memory \citep{wang2023voyager, zhai2025agentevolver, lin2025se, sun2025seagent, wang2026atlasva}. In realistic tool-use settings with multi-turn interfaces, strict execution constraints, and partial failures \citep{li2023api, xie2024osworld, liu2025agent, yin2026glove, peng2026tool, yang2026evotool}, improvement depends not only on how the policy is optimized but also on what learning signal the training-time environment exposes while the policy changes.

Existing self-evolving agents emphasize different parts of this loop. \emph{Model-centric} methods improve the policy through rollout replay, self-reflection, reward optimization, or post-training \citep{huang2025r, yuan2025agent, wang2025ragen, xiang2026systematic}, but against a fixed environment, so the learning signal is a function of the policy's own rollouts; with sparse rewards and long horizons, this self-referential signal produces exploration bias, unstable recovery, and inefficient credit assignment \citep{shridhar2020alfworld}. \emph{Environment-centric} methods instead adapt curricula, task distributions, synthetic instructions, or interaction experiences \citep{bengio2009curriculum, lu2026dont, yang2026coevolve, hao2026failure}, including methods that use rollout or failure feedback. Our focus is to couple training-interface adaptation and policy reweighting through a shared executable diagnosis. We study a training-time gap we call \textbf{Agent--Learning Environment Misalignment (ALEM)}: \emph{as the agent's capability frontier shifts during training, the learning environment that supplies interaction support and supervision does not track newly revealed failures, so the signal stays static, weakly targeted, or uninformative}. Throughout, ``environment'' means the training-time learning substrate (task exposure, observation interfaces, action constraints, recovery feedback), never the evaluation benchmark, tool semantics, or verifier. Closing the loop raises two challenges, both visible in the flight example. \cone\ \emph{Uninformative feedback}: the environment tells the agent \emph{that} a call failed, not \emph{why}, so the specific error cannot be repaired. \ctwo\ \emph{Undifferentiated credit}: a terminal scalar reward gives every failed rollout the same zero, so optimization pressure cannot be aimed at failures that are attributable and recoverable (a wrong argument type) rather than at failures that only surface in the final response check.

\begin{figure*}[t]
    \centering
    \vspace{-1.0em}
    \includegraphics[width=1\textwidth,trim=0 0.0cm 0 0.0cm,clip]{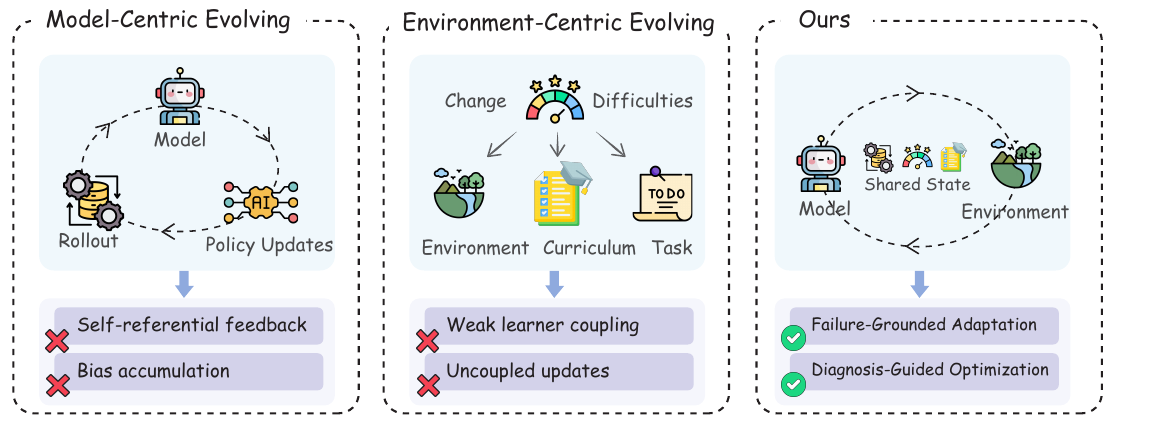}
    \vspace{-1.0em}
    \caption{\small
        \textbf{Three ways to self-evolve a tool-use agent.}
        Left and middle illustrate model-centric and environment-centric designs with uncoupled updates. \ours\ (right) explicitly couples complementary agent-side and learning-environment-side adaptations through a shared verifier-grounded diagnosis, with tasks, tools, rewards, and the verifier unchanged.
    }
        \vspace{-1.0em}
    \label{fig:seal_overview}
\end{figure*}

We propose \textbf{\ours}, a closed-loop co-evolution framework for multi-turn tool-use agents (Figure~\ref{fig:seal_overview}). Here, \emph{synergistic} denotes the coupled design: shared diagnosis coordinates the current policy update and subsequent interaction support; Section~\ref{sec:ablation} examines their empirical complementarity. \ours\ diagnoses every rollout with executable evidence (parser, schema, execution, state, and verifier checks) into turn-level labels such as a missing call or a state mismatch. For \cone, the training-time interface evolves to expose schema affordances, constraints, and recovery-oriented feedback selected by the agent's recurring failures. For \ctwo, policy optimization reweights each trajectory's GRPO advantage by how attributable and recoverable its failures are. The environment thus adapts to the current policy's failures, not a difficulty schedule. Crucially, \ours\ never touches the benchmark: tasks, tools, rewards, and the verifier stay fixed; the interface only re-exposes existing information, never reference calls, hidden parameters, or answers; and every training-only addition is removed at evaluation, so \ours\ and its baselines are tested under identical conditions. This conservatism lets a gain be attributed to a better policy rather than a friendlier test. Because diagnosis needs executable evidence, we study interactive tool use, where such evidence is structured and auditable. Our contributions are threefold:
\begin{itemize}[topsep=-1pt,itemsep=1.5pt,parsep=0pt,partopsep=0pt]
    \item We study \textbf{ALEM}, the training-time mismatch between an evolving agent's execution failures and its learning environment, and trace it to two challenges: uninformative feedback (\cone) and undifferentiated credit (\ctwo).
    \item We introduce \textbf{\ours}, where a shared verifier-grounded diagnosis coordinates learning-interface evolution and diagnosis-guided advantage reweighting, while keeping tasks, tools, rewards, and the verifier fixed.
    \item We show that \ours\ improves over matched Vanilla RL across three backbones (+9.50 points at 7B), exceeds Environment Tuning in reported point estimates, and transfers positively to the evaluated held-out domains; a matched factorial ablation yields a +5.00-point interaction estimate.
\end{itemize}

\section{Related Work}
\label{sec:related}

\stitle{One-sided self-evolution.}
Model-side methods refine the agent through skill learning, self-consolidation, prompt adaptation, memory, or RL from interaction feedback \citep{xia2026skillrl, yu2026self, agrawal2025gepa, zweiger2025selfadapting}: experience is internalized into the policy, prompt, memory, or skills while the training-time environment stays fixed, so the feedback is only as informative as that environment makes it. Environment-side methods instead reshape what the learner sees through curricula, synthetic instructions, task evolution, or tool and skill construction \citep{matiisen2019teacher, portelas2020automatic, wang2023selfinstruct, xu2024evolinstruct, cai2023large, qian2023creator, yuan2023craft}, acting on task diversity, difficulty, or coverage, driven by task-level success rather than by a diagnosis of why the current policy fails. In this work, we keep the model-side update and make environment adaptation \emph{failure-conditioned}: verifier-grounded diagnoses decide which cues and recovery feedback the current policy sees.

\stitle{Environments and closest methods.}
Interactive benchmarks for tool use, web navigation, operating-system control, embodied simulation, and software engineering expose the multi-turn dependencies, sparse rewards, and recovery dynamics \ours\ targets \citep{patil2025berkeley, li2023api, zhou2023webarena, liu2023agentbench, jimenez2023swe}; recent work co-evolves agents with generated or edited environments \citep{xiagentgym, zhang2025autoenv, guo2025genenv}, whereas \ours\ keeps tasks, tool semantics, and the verifier fixed and co-evolves only the training-time interface. ALIGN \citep{liu2025agent} augments interfaces for agent--environment alignment; \ours\ instead uses verifier-grounded diagnoses to adjust training-time support and policy updates, addressing ALEM. Closest in mechanism, Dynamic Cheatsheet \citep{suzgun2025dynamiccheatsheet} evolves an inference-time memory without updating parameters, whereas \ours's guidance exists only at training time; Fission-GRPO \citep{zhang2026fissiongrpo} turns failures into corrective instances with a learned error simulator and extra rollouts, whereas \ours\ reweights existing rollouts with deterministic diagnosis; and Environment Tuning \citep{lu2026dont}, our closest method-level comparison (Table~\ref{tab:bfcl_v3_results}), adapts task exposure, feedback, and progress supervision, whereas \ours\ keeps the binary verifier reward and changes only the interface and optimization weights.

\section{Methodology}
\label{sec:method}

\begin{figure*}[t]
    \centering
    \includegraphics[width=0.9\textwidth,trim=0.10cm 0.05cm 0.15cm 0.03cm,clip]{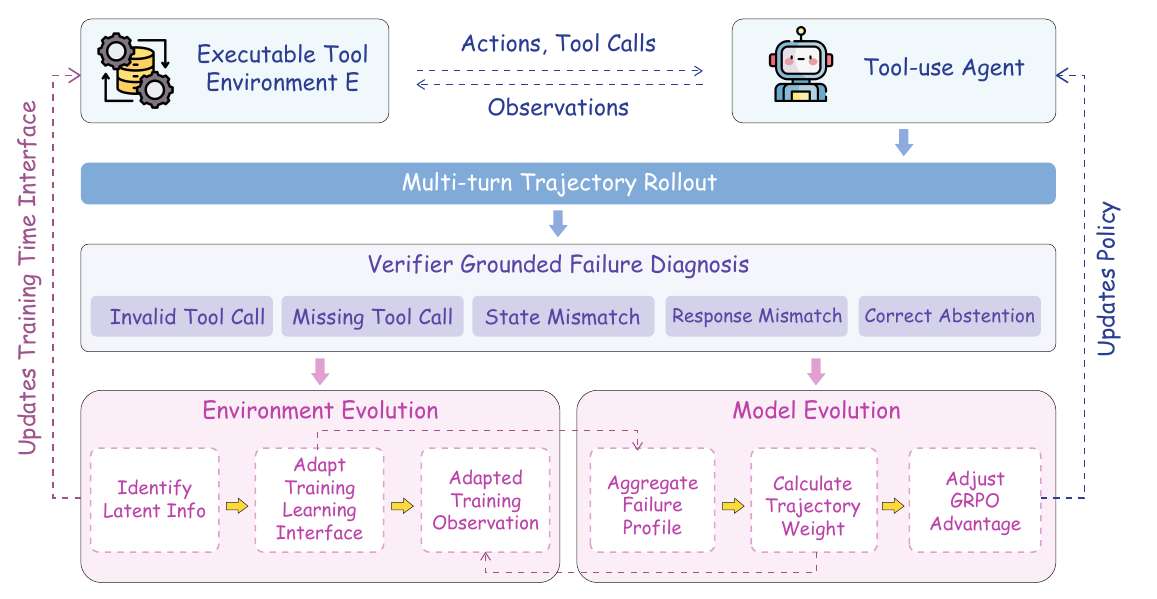}
    \vspace{-0.8em}
    \caption{\small
    \textbf{The \ours\ training loop.}
    The agent interacts with the executable tool environment to produce
    multi-turn trajectories (top), which executable evidence and the
    verifier diagnose into turn-level failure labels (middle). The same
    labels drive environment evolution (bottom left: identify the missing
    information, adapt the training-time interface, expose an adapted
    observation in the next round) and model evolution (bottom right:
    aggregate the failure profile, compute trajectory weights, rescale the
    GRPO advantages of the current update), while tool semantics, rewards,
    and evaluation never change. The diagnosis block shows behavioral
    categories; the operational labels are in Table~\ref{tab:diagnosis_labels}.
}
    \label{fig:seal_training_loop}
    \vspace{-1.0em}
\end{figure*}

Figure~\ref{fig:seal_training_loop} gives an overview of \ours. Standard RL gives a tool-use agent one binary reward per episode; \ours\ keeps that reward but, during training, also \textbf{diagnoses every rollout turn by turn from executable evidence}. These diagnoses are used twice: they decide \emph{which feedback} the training-time interface adds for the agent, such as recovery hints, and \emph{how much} each trajectory counts in the policy update. We take the view of a trainer who controls training but not evaluation, and separate the \emph{tool backend}, which defines what a call does, from the \emph{learning interface}, which defines what the agent observes. \ours\ changes \textbf{only the learning interface}: tools, task labels, rewards, and the verifier never change, so the benchmark still measures the same thing.

\subsection{Problem Formulation}
\label{sec:formulation}

We model interactive tool use as a partially observable decision process with hidden states, horizon $H$, and a tool set $\mathcal{F}$: actions include natural-language responses and executable tool calls, and observations include dialogue context, tool outputs, and execution errors. Given an instruction $q\sim\mathcal{D}$, a policy $\pi_\theta$ interacting with an executable environment $\mathcal{E}$ induces a rollout
\begin{equation}
\tau=\{(u_i,a_i,o_i)\}_{i=1}^{T}\sim P_{\mathcal{E}}(\cdot\mid q,\pi_\theta),
\qquad T\leq H,
\end{equation}
where $u_i$ is the dialogue context, $a_i$ the model action, and $o_i$ the environment observation. The executable verifier $\mathcal{V}$ returns a terminal binary reward $r(\tau)=\mathcal{V}(\tau,q,\mathcal{E})\in\{0,1\}$, and standard RL maximizes
$J(\pi_\theta)=\mathbb{E}_{q\sim\mathcal{D},\,\tau\sim P_{\mathcal{E}}(\cdot\mid q,\pi_\theta)}[r(\tau)]$.
This scalar says whether a trajectory succeeded, not why it failed. \ours\ therefore augments the training-time feedback to $(r(\tau),Z(\tau))$, where $Z(\tau)$ is a structured failure diagnosis extracted from the executable interaction trace; the reward and evaluation protocol are unchanged.

\subsection{Verifier-Grounded Failure Diagnosis}
\label{sec:failure-diagnosis}

For each rollout, \ours\ produces turn-level labels $Z(\tau)=\{z_1,\ldots,z_T\}$, $z_i\in\mathcal{Z}$, where $z_i$ is the dominant outcome or failure mode at turn $i$:
\begin{equation}
z_i=\Psi(a_i,\xi_i,s_{i-1},s_i,\mathcal{F}),
\end{equation}
with $\xi_i$ the executable evidence at turn $i$ and $s_{i-1},s_i$ the pre- and post-action states, when available.

\stitle{Rules, not a judge.} $\Psi$ is a deterministic classifier over executable traces (parser checks, schema validation, execution errors, state transitions, verifier comparisons), applied in a fixed priority order that puts directly executable failures before downstream verifier-level ones (Appendix~\ref{app:diagnosis_priority}). We use rules rather than an LLM judge because executable checks are trace-grounded and reproducible, whereas a learned judge can reintroduce the self-referential feedback of model-centric evolution.

\stitle{Seven labels.} Table~\ref{tab:diagnosis_labels} lists the labels, their executable conditions (checked top to bottom), and their base utilities $\rho_0$. \texttt{response\_mismatch} is the BFCL execution-response check, not the wording of a final answer, and bookkeeping outcomes (\texttt{checker\_error}, \texttt{gt\_error}) are excluded from weighting and recovery selection.

\begin{table}[t]
\centering
\caption{\small
\textbf{Operational diagnosis labels.} $\Psi$ evaluates the conditions in order and assigns the first that holds; acceptance is defined by the original verifier. $\rho_0$ is the base utility of Section~\ref{sec:advantage-reweighting}: labels with concrete executable evidence and a clear repair direction weigh more than downstream verifier-level mismatches.
}
\label{tab:diagnosis_labels}
\small
\renewcommand{\arraystretch}{1.0}
\setlength{\tabcolsep}{4pt}
\begin{tabularx}{\linewidth}{@{}r l X c@{}}
\toprule
Order & Label & Executable condition & $\rho_0$ \\
\midrule
1a & \texttt{correct\_abstention} & No tool call is required, and the model makes none. & 1.8 \\
1b & \texttt{spurious\_tool\_call} & No tool call is required, but the model attempts one. & 2.0 \\
2 & \texttt{empty\_turn\_model\_response} & A tool call is required, but none is produced. & 1.5 \\
3 & \texttt{instance\_mismatch} & Model-side and reference-side executor object sets differ. & 0.6 \\
4 & \texttt{state\_mismatch} & The official state checker rejects the resulting state. & 1.0 \\
5 & \texttt{response\_mismatch} & State passes, but the execution-response checker rejects. & 0.9 \\
6 & \texttt{pass} & Both state and execution-response checks pass. & 1.0 \\
\bottomrule
\end{tabularx}
\end{table}

\stitle{Reliability.} Deterministic rules can still mislabel, so we audited them: in a blinded audit of 60 label-stratified diagnoses (Appendix~\ref{app:diagnosis_audit}), the annotator agreed with $\Psi$ on the exact label in 57 cases (95.0\%) and on failure versus non-failure in 59, with every disagreement on state versus execution-response matching. Diagnosis never touches the reward: a failed trajectory still receives $r(\tau)=0$.

\subsection{Learning-Interface Evolution}
\label{sec:interface-evolution}

\ours\ evolves only the \emph{training-time learning interface}, never the verifier, tool signatures, tool outputs, or task labels. At turn $i$ it augments the original observation $o_i$ as
\begin{equation}
\tilde{o}_i=\Omega_t(o_i,\mathcal{F},H_t,C_{t-1})
=o_i\oplus\phi_{\mathrm{schema}}(\mathcal{F})\oplus\phi_{\mathrm{err}}(o_i,H_t)\oplus\phi_{\mathrm{cap}}(C_{t-1}),
\end{equation}
where $H_t$ comes from previous rollouts and $C_{t-1}$ is the latest completed failure profile (empty at $t=0$); this round's $C_t$ affects $\Omega_{t+1}$ only after sampling. Each component is a deterministic transformation or fixed template, not an LLM:
\begin{itemize}[topsep=1pt,itemsep=1pt,parsep=0pt,partopsep=0pt,leftmargin=1.2em]
    \item \textbf{Schema cues} ($\phi_{\mathrm{schema}}$) make schema constraints explicit (required arguments, enums, types, call formats), because most invalid calls violate a constraint the schema already states.
    \item \textbf{Recovery feedback} ($\phi_{\mathrm{err}}$) turns a generic execution error into a hint on how to recover, without revealing the answer, because a bare error is the uninformative feedback of Section~\ref{sec:intro}.
    \item \textbf{Capability cues} ($\phi_{\mathrm{cap}}$) add a class-level instruction for each recurring failure: cautious calls for \texttt{spurious\_tool\_call}, an active response for \texttt{empty\_turn\_model\_response}, and state-aware recovery for \texttt{state\_mismatch} (templates in Appendix~\ref{app:prompt_templates}).
\end{itemize}
\stitle{Safeguards.} The cues describe how to repair an error class, never the reference calls, hidden parameters, or final answer, and the whole interface is removed at evaluation. Proposition~\ref{prop:interface} (Appendix~\ref{app:theory}) shows that, for unrestricted policies and append-only observations, accepted trajectories and optimal-policy availability are preserved. For a finite set $\mathcal{L}$ of complete template configurations, the additional single-round signal carries at most $\log_2|\mathcal{L}|$ bits about the reference conditioned on raw observations and schema; this does not guarantee absence of leakage.

\subsection{Diagnosis-Guided Advantage Reweighting}
\label{sec:advantage-reweighting}

Under a binary reward, GRPO~\citep{shao2024deepseekmath} gives every failed trajectory in a group the same advantage, $-\sqrt{k/(G-k)}$ for $k$ successes among $G$ rollouts (Lemma~\ref{lem:credit}, Appendix~\ref{app:theory}), so an invalid argument and a wrong final answer are pushed down equally. \ours\ instead weights each trajectory by what its diagnoses reveal, in three steps.

\stitle{Diagnostic profile.} For trajectory $\tau_j$ with valid, non-bookkeeping turn labels $\mathcal{I}_j$, the profile $p_j(z)=\frac{1}{|\mathcal{I}_j|}\sum_{i\in\mathcal{I}_j}\mathbf{1}[z_i=z]$ is the share of its turns carrying label $z\in\mathcal{Z}$.

\stitle{Label utilities.} A utility $\rho_t(z)>0$ scores how actionable and attributable label $z$ is. The fixed base utilities $\rho_0$ (Table~\ref{tab:diagnosis_labels}) favor failures with concrete executable evidence and a clear repair, such as \texttt{spurious\_tool\_call}, over ambiguous ones such as \texttt{response\_mismatch}; \texttt{pass} stays at 1.0. For a failure label $z$, let $n_t(z)$ count its occurrences in the statistics window, $p_t(z)$ be its share of non-bookkeeping occurrences, $\bar r_t(z)$ the mean verifier reward of the trajectories containing it, and $c_t(z)=\min(1,n_t(z)/32)$ a support factor. The realized utilities then adapt as
\begin{align}
\rho_t^*(z)&=\operatorname{clip}_{[0.5,2.0]}
\left[\rho_0(z)\{1+p_t(z)(1-\bar r_t(z))c_t(z)\}\right],
\label{eq:utility_target}\\
\rho_t(z)&=0.75\,\rho_{t-1}(z)+0.25\,\rho_t^*(z),
\label{eq:utility_update}
\end{align}
so a label gains weight while it is frequent and rarely co-occurs with success, while low-support labels exert smaller utility updates. By Proposition~\ref{prop:utility}, utilities stay in $[0.5,2.0]$ and move by at most 0.375 per round; when a failure type disappears, the EMA gradually relaxes its utility toward $\rho_0$. All constants are fixed by the comparison protocol.

\stitle{Weighted advantage.} Each trajectory's weight averages the utilities over its profile,
\begin{equation}
w_j=\operatorname{clip}\Big(\sum_{z\in\mathcal{Z}}p_j(z)\,\rho_t(z),\;w_{\min},\;w_{\max}\Big),
\qquad [w_{\min},w_{\max}]=[0.5,2.0],
\end{equation}
with $w_j=1$ when no valid label exists (the clip never binds, Corollary~\ref{cor:convex}). \ours\ scales the group-normalized GRPO advantage to $\widetilde{A}_j=w_jA_j$ on all trainable response tokens, for positive and negative advantages alike, without scaling the reward, adding token-level weights, or renormalizing. Because $w_j>0$, weighting preserves each group-relative advantage's sign and no sampled trajectory is discarded. The weighted reward-driven score-function term admits a tilted-distribution representation with a bounded normalization factor and density ratio to $\pi_\theta$ in $[1/4,4]$ (Proposition~\ref{prop:tilt}); this does not equate the full clipped, KL-regularized GRPO update to an unweighted one.

\subsection{Training Loop}
\label{sec:algorithm}

Each round is one optimizer iteration over 32 prompts with eight rollouts each (Algorithm~\ref{alg:seal}, Appendix~\ref{app:algorithm}) and runs four steps: (1)~\textbf{collect} trajectories $\{\tau_j\}_{j=1}^{N}$ by running $\pi_{\theta_t}$ under interface $\Omega_t$; (2)~\textbf{diagnose} each $\tau_j$ into its verifier reward $r(\tau_j)$ and labels $Z(\tau_j)$, and aggregate the failure profile $C_t=\operatorname{Agg}(\{Z(\tau_j)\})$; (3)~\textbf{update the policy} on this batch, $\theta_{t+1}=\operatorname{GRPO}(\theta_t;\{\tau_j,r(\tau_j),w_jA_j\})$; and (4)~\textbf{evolve the interface} for the next round, $\Omega_{t+1}=\operatorname{Evolve}(\Omega_t,C_t)$. Diagnoses never alter trajectories already sampled, and tools, task labels, rewards, and the verifier stay fixed throughout. The two sides also cover each other's blind spot: reweighting redistributes credit only within groups that contain both successes and failures, whereas the interface changes the rollouts of prompts the policy never solves (Remark~\ref{rem:division}, Appendix~\ref{app:theory}).

\section{Experiments}
\label{sec:exp}

We test, through four questions, our central claim: a tool-use agent learns more from the same training data when it co-evolves with its learning environment. \textbf{RQ1}: Does \ours\ improve in-distribution multi-turn tool use over RL with the same training examples and rollout budget (Section~\ref{sec:main_results})? \textbf{RQ2}: Do the gains transfer to held-out tool domains (Section~\ref{sec:ood_results})? \textbf{RQ3}: Do the two diagnosis-driven adaptations provide complementary gains when coupled (Section~\ref{sec:ablation})? \textbf{RQ4}: How do the gains emerge during training (Section~\ref{sec:training_dynamics})?

\subsection{Experimental Setup}
\label{sec:exp_settings}

\stitle{Benchmarks.}
In-distribution: the BFCL V3 multi-turn subset \citep{patil2025berkeley}, 800 examples in four categories (\textit{Base}, \textit{Missing Functions}, \textit{Missing Parameters}, \textit{Long Context}), split at the scenario-ID level (seed 42) into 400 training and 400 task-disjoint evaluation examples, 100 per category (Appendix~\ref{app:benchmark_details}). Out-of-distribution: the BFCL V4 Web Search and Memory tracks and the Retail, Airline, and Telecom domains of $\tau^2$-bench~\citep{barres2025tau2}, which differ from BFCL V3 in tool domains, schemas, and interaction patterns. All evaluations use the original benchmark environments and verifiers.

\stitle{Backbones.}
We train Qwen2.5-3B-Instruct and Qwen2.5-7B-Instruct \citep{yang2024qwen25}, general instruction-tuned agents, and ToolACE-2-Llama-3.1-8B \citep{liu2025toolace}, a stronger tool-specialized initialization that tests whether the method still helps a model that already calls tools well. The proprietary and open-source reference systems in Tables~\ref{tab:bfcl_v3_results} and~\ref{tab:ood_results} are described, with references, in Appendix~\ref{app:models}.

\stitle{Conditions.}
All binary-reward conditions share the backbone, the 400-example split, eight rollouts per prompt, GRPO and optimizer settings, and the binary verifier reward, and are evaluated with every training-only addition removed. \textsc{Vanilla RL} is GRPO with the original interface (the matched training baseline). \ours\ adds training-interface enrichment and diagnosis-guided advantage reweighting; the component settings are described in Appendix~\ref{app:matched_controls}.
Environment Tuning \citep{lu2026dont}, the closest published method (curriculum, hand-written error hints, per-turn progress reward), is trained on the same split with its released recipe; it is a method-level rather than reward-matched comparison.
Optimizer settings are listed in Table~\ref{tab:hyperparameters}. The training comparisons report one run per condition. The main Qwen2.5-7B SEAL checkpoint uses the complete online configuration: fixed schema affordances, diagnosis-conditioned recovery selection, and diagnosis-aware advantage reweighting. Diagnoses reweight the current policy update and select recovery guidance for subsequent training rounds; all adaptive guidance is removed at evaluation (Appendix~\ref{app:run_correspondence}).

\begin{table*}[!t]
    \centering
    \caption{\small
        \textbf{Main results on the BFCL V3 multi-turn benchmark (accuracy, \%).}
        Our training comparisons use 400 examples and report one run per condition. The Qwen2.5-7B SEAL row uses the complete online configuration (Appendix~\ref{app:run_correspondence}). Parentheses give gains over the untrained backbone. Reference systems use different scales, data, and recipes (Appendix~\ref{app:models}).
    }
    \label{tab:bfcl_v3_results}
    \small\setlength{\tabcolsep}{4pt}
    \resizebox{\textwidth}{!}{%
        \begin{tabular}{lccccc}
            \toprule
            \multirow{2}{*}{\textbf{Model}} & \multicolumn{5}{c}{\textbf{BFCL V3 Multi-Turn}} \\
            \cmidrule(lr){2-6}
            & \textbf{Average} & \textbf{Base} & \textbf{Miss Func} & \textbf{Miss Param} & \textbf{Long Context} \\
            \midrule

            \multicolumn{6}{l}{\textbf{Closed-source}} \\
            Gemini-3-Pro-Preview
            & 60.75 & 64.50 & 60.00 & 54.50 & 64.00 \\
            Claude-Sonnet-4.5
            & 61.38 & 69.00 & 65.00 & 52.50 & 59.00 \\
            GPT-4o-2024-11-20
            & 42.50 & 55.50 & 34.50 & 29.00 & 51.00 \\
            Gemini-2.5-Flash
            & 36.25 & 41.50 & 36.00 & 32.00 & 35.50 \\
            \midrule

            \multicolumn{6}{l}{\textbf{Open-source}} \\
            GLM-4.6 355B
            & 68.00 & 74.50 & 68.00 & 63.00 & 66.50 \\
            Qwen3-32B
            & 47.88 & 56.00 & 52.50 & 40.00 & 43.00 \\
            Qwen2.5-14B-Instruct
            & 25.25 & 33.00 & 26.00 & 22.00 & 20.00 \\
            xLAM-2-3b-fc-r
            & 58.38 & 71.50 & 59.00 & 57.50 & 45.50 \\
            \midrule

            Qwen2.5-3B-Instruct
            & 5.75 & 11.00 & 6.00 & 3.00 & 3.00 \\
            \rowcolor{teal!10}
            \, + \textsc{Vanilla RL}
            & 9.25 \improvement{(+3.50)} & 16.00 \improvement{(+5.00)} & 9.00 \improvement{(+3.00)} & 6.00 \improvement{(+3.00)} & 6.00 \improvement{(+3.00)} \\
            \rowcolor{gray!8}
            \, + Environment Tuning
            & 11.50 \improvement{(+5.75)} & 16.00 \improvement{(+5.00)} & 10.00 \improvement{(+4.00)} & 8.00 \improvement{(+5.00)} & 12.00 \improvement{(+9.00)} \\
            \rowcolor{orange!10}
            \, + \ours
            & 14.00 \improvement{(+8.25)} & 19.00 \improvement{(+8.00)} & 15.00 \improvement{(+9.00)} & 12.00 \improvement{(+9.00)} & 10.00 \improvement{(+7.00)} \\
            \midrule

            Qwen2.5-7B-Instruct
            & 14.00 & 22.00 & 14.00 & 10.00 & 10.00 \\
            \rowcolor{teal!10}
            \, + \textsc{Vanilla RL}
            & 30.75 \improvement{(+16.75)} & 46.00 \improvement{(+24.00)} & 27.00 \improvement{(+13.00)} & 27.00 \improvement{(+17.00)} & 23.00 \improvement{(+13.00)} \\
            \rowcolor{gray!8}
            \, + Environment Tuning
            & 37.00 \improvement{(+23.00)} & 52.00 \improvement{(+30.00)} & 34.00 \improvement{(+20.00)} & 35.00 \improvement{(+25.00)} & 27.00 \improvement{(+17.00)} \\
            \rowcolor{orange!10}
            \, + \ours
            & 40.25 \improvement{(+26.25)} & 58.00 \improvement{(+36.00)} & 36.00 \improvement{(+22.00)} & 34.00 \improvement{(+24.00)} & 33.00 \improvement{(+23.00)} \\
            \midrule

            ToolACE-2-Llama-3.1-8B
            & 32.00  & 45.00  & 26.00  & 35.00  & 22.00 \\
            \rowcolor{teal!10}
            \, + \textsc{Vanilla RL}
            & 38.50 \improvement{(+6.50)}
            & 52.00 \improvement{(+7.00)}
            & 30.00 \improvement{(+4.00)}
            & 40.00 \improvement{(+5.00)}
            & 32.00 \improvement{(+10.00)} \\
            \rowcolor{gray!8}
            \, + Environment Tuning
            & 43.25 \improvement{(+11.25)}
            & 54.00 \improvement{(+9.00)}
            & 41.00 \improvement{(+15.00)}
            & 36.00 \improvement{(+1.00)}
            & 42.00 \improvement{(+20.00)} \\
            \rowcolor{orange!10}
            \, + \ours
            & 46.75 \improvement{(+14.75)}
            & 58.00 \improvement{(+13.00)}
            & 46.00 \improvement{(+20.00)}
            & 44.00 \improvement{(+9.00)}
            & 39.00 \improvement{(+17.00)} \\

            \bottomrule
        \end{tabular}%
    }
\end{table*}

\subsection{In-Distribution Results on BFCL V3 (RQ1)}
\label{sec:main_results}

\stitle{\ours\ beats RL on every reported backbone and category, with higher point estimates than the closest environment-side method.}
Table~\ref{tab:bfcl_v3_results} supports this from five angles. \emph{Consistency}: under the same rollouts, optimizer, and reward, \ours\ exceeds \textsc{Vanilla RL} in all 12 backbone-by-category cells, so the gain is not carried by a favorable subset. \emph{Magnitude}: its total improvement over the untrained checkpoint is 1.6 to 2.4 times that of RL alone, reaching +26.25 points on Qwen2.5-7B, which suggests that a sparse terminal reward leaves useful learning signal unused. \emph{Initialization}: the tool-specialized ToolACE-2 also gains substantially over its RL counterpart, suggesting that diagnosis-guided training complements tool-specific pre-training. \emph{Failure type}: averaged across backbones relative to Vanilla RL, \textit{Missing Functions} improves most, \textit{Long Context} ties \textit{Base}, and \textit{Missing Parameters} improves least, plausibly because recognizing an underspecified argument depends on the user's instruction rather than the tool schema. \emph{Method-level baseline}: \ours\ has higher point estimates than Environment Tuning on every backbone (by 2.50 to 3.50 points) and in 9 of 12 cells while needing none of its curriculum, hand-written per-domain hints, or dense progress reward; \ours\ keeps the binary reward and derives its cues from diagnosed failure classes. Against the reference rows, 400 examples take Qwen2.5-7B past Gemini-2.5-Flash and its own 14B sibling, and ToolACE-2 with \ours\ past GPT-4o to within about one point of the four-times larger Qwen3-32B, though the strongest systems remain well ahead. In short, diagnosis-driven co-evolution recovers learning signal that RL alone leaves on the table, across scale, initialization, and failure type; we next ask whether the gains are specific to BFCL V3.

\subsection{Out-of-Distribution Transfer (RQ2)}
\label{sec:ood_results}

\stitle{\ours\ shows positive transfer in the evaluated held-out tool domains.}
Table~\ref{tab:ood_results} evaluates the BFCL~V3-trained Qwen2.5 models on BFCL V4 and $\tau^2$-bench, whose tools never appear in training, and four angles show that the gains carry over. \emph{Comparison}: \ours's gains on the held-out averages are 2.2 to 4.3 times those of \textsc{Vanilla RL}, which moves them by about one point. \emph{Breadth}: \ours\ improves all ten backbone-by-track cells over the untrained model and matches or beats \textsc{Vanilla RL} in each, even on Web Search, whose search tools have no counterpart in BFCL V3. \emph{Magnitude}: the largest gain comes in a domain absent from training, $\tau^2$-bench Retail, where Qwen2.5-7B rises from 16.67\% to 26.32\%, more than three times the gain of RL alone. \emph{Mechanism}: \ours's cues and credit target failure classes (for instance, changing state without checking it first) rather than BFCL tasks, and the cues are removed at evaluation; these scores provide preliminary evidence that BFCL V3 gains transfer to the evaluated held-out domains. Absolute scores are still low, and Web Search stays near zero at 3B, so cross-benchmark generalization remains hard. In short, the observed gains extend to the evaluated held-out benchmarks; we next ask which parts of the system are responsible.

\begin{table*}[!t]
    \centering
    \caption{\small
        \textbf{Out-of-distribution transfer to BFCL V4 and $\tau^2$-bench (accuracy, \%).}
        SEAL and Vanilla RL reuse the Table~\ref{tab:bfcl_v3_results} checkpoints without OOD-specific training or selection; each was trained on 400 BFCL V3 examples and reports one run. Parentheses give gains over the untrained backbone. Reference rows are not matched training controls (Appendix~\ref{app:models}).
    }
    \label{tab:ood_results}
    \small\setlength{\tabcolsep}{4pt}
    \resizebox{\textwidth}{!}{
        \begin{tabular}{l cccccccc}
            \toprule
            \multirow{2}{*}{\textbf{Model}}
            & \multicolumn{3}{c}{\textbf{BFCL V4}}
            & \multicolumn{4}{c}{\textbf{$\tau^2$-bench}} \\
            \cmidrule(lr){2-4} \cmidrule(lr){5-8}
            & \textbf{Avg.}
            & \textbf{Web Search}
            & \textbf{Memory}
            & \textbf{Avg.}
            & \textbf{Retail}
            & \textbf{Airline}
            & \textbf{Telecom} \\
            \midrule

            BitAgent-8B
            & 7.41
            & 4.50
            & 10.32
            & 10.00
            & 7.50
            & 15.00
            & 7.50 \\

            ToolACE-2-Llama-3.1-8B
            & 15.90
            & 9.00
            & 22.80
            & 10.83
            & 10.00
            & 15.00
            & 7.50 \\

            watt-tool-8B
            & 8.67
            & 4.00
            & 13.33
            & 13.33
            & 15.00
            & 20.00
            & 5.00 \\

            Qwen3-30B-A3B-Instruct
            & 20.07
            & 22.50
            & 17.63
            & 32.03
            & 50.88
            & 32.50
            & 12.72 \\

            \midrule

            Qwen2.5-3B-Instruct
            &  4.69
            &  0.00
            &  9.38
            &  9.64
            &  5.26
            &  14.00
            &  9.65        \\
            \rowcolor{teal!10}
            \, + \textsc{Vanilla RL}
            & 5.88 \improvement{(+1.19)}
            & 0.50
            & 11.26
            & 10.60 \improvement{(+0.96)}
            & 6.14
            & 16.00
            & 9.65    \\

            \rowcolor{orange!10}
            \, + \ours
            & 8.63 \improvement{(+3.94)}
            & 2.00
            & 15.26
            & 11.76 \improvement{(+2.12)}
            & 8.77
            & 16.00
            & 10.52    \\

            \midrule

            Qwen2.5-7B-Instruct
            &  9.17
            &  6.50
            &  11.83
            &  16.11
            &  16.67
            &  20.00
            &  11.67        \\
            \rowcolor{teal!10}
            \, + \textsc{Vanilla RL}
            & 10.26 \improvement{(+1.09)}
            & 7.50
            & 13.02
            & 17.19 \improvement{(+1.08)}
            & 19.30
            & 20.00
            & 12.28   \\

            \rowcolor{orange!10}
            \, + \ours
            & 12.71 \improvement{(+3.54)}
            & 9.50
            & 15.92
            & 20.79 \improvement{(+4.68)}
            & 26.32
            & 22.00
            & 14.04      \\

            \bottomrule
        \end{tabular}
    }
\end{table*}

\subsection{Ablation Study (RQ3)}
\label{sec:ablation}

\stitle{In this run, the two sides work best together.}
Table~\ref{tab:seal_ablation} crosses interface selection and advantage weighting on Qwen2.5-7B. \emph{Hierarchy}: replacing diagnosis-aware weighting costs 5.75 points and replacing dynamic interface selection costs 4.75, yet every variant still beats \textsc{Vanilla RL}. \emph{Complementarity}: relative to Static+Uniform, diagnosis-aware weighting alone changes accuracy by +0.75 points and dynamic interface selection alone by $-0.25$; together they add 5.50, yielding a positive interaction point estimate of +5.00 points. This is consistent with Remark~\ref{rem:division}: the interface can turn unsolved prompts into the mixed success-failure groups on which reweighting acts. \emph{Adaptivity}: fixing the interface and using uniform weighting gives up 5.50 points of the full configuration's accuracy. \emph{Consistency}: full \ours\ is best in every category, and no variant falls below \textsc{Vanilla RL} in any. In this single-run comparison, the pattern is consistent with complementary adaptations coordinated through a shared execution diagnosis.

\begin{table*}[t]
    \centering
    \caption{\small
        \textbf{Factorial ablation of interface adaptation and diagnosis-aware optimization.}
        Qwen2.5-7B, BFCL V3 multi-turn accuracy (\%). Each condition reports one run;
        the interaction is a point estimate. Full SEAL uses Table~\ref{tab:bfcl_v3_results}'s checkpoint.
        $\Delta$ gives differences from Full SEAL in percentage points.
    }
    \label{tab:seal_ablation}
    \small
    \setlength{\tabcolsep}{4pt}
    \renewcommand{\arraystretch}{1.12}
    \resizebox{\textwidth}{!}{%
    \begin{tabular}{llcccccc}
        \toprule
        \textbf{Interface} & \textbf{Weighting} & \textbf{Avg.} & $\boldsymbol{\Delta}$
        & \textbf{Base} & \textbf{Miss Func} & \textbf{Miss Param} & \textbf{Long Context} \\
        \midrule
        \rowcolor{tablegray}
        Static & Uniform & 34.75 & $-5.50$ & 52.00 & 32.00 & 32.00 & 23.00 \\
        Dynamic & Uniform & 34.50 & $-5.75$ & 49.00 & 34.00 & 31.00 & 24.00 \\
        Static & Diagnosis-aware & 35.50 & $-4.75$ & 50.00 & 35.00 & 32.00 & 25.00 \\
        \rowcolor{tableaccent}
        Dynamic & Diagnosis-aware & \textbf{40.25} & --- & 58.00 & 36.00 & 34.00 & 33.00 \\
        \bottomrule
    \end{tabular}%
    }
\end{table*}
\FloatBarrier

\subsection{Training Dynamics (RQ4)}
\label{sec:training_dynamics}

\begin{figure*}[!t]
    \centering
    \includegraphics[width=1.00\textwidth,trim=0.05cm 0.18cm 0.10cm 0.18cm,clip]{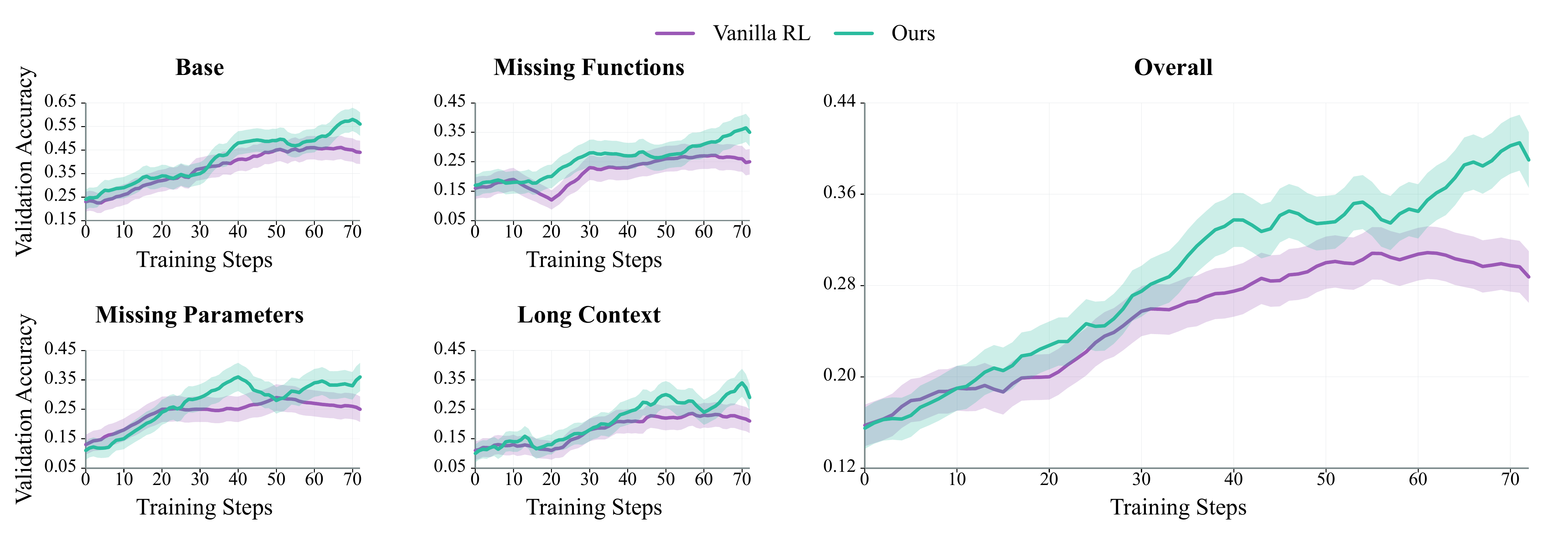}
    \vspace{-0.8em}
    \caption{\small
        \textbf{Training dynamics of \textsc{Vanilla RL} and \ours\ on the BFCL V3 validation subsets.}
        Validation accuracy across training steps per category and overall. Shading is the recorded binomial approximation, not training-seed variation or an interval for paired differences; the overall band does not account for dependence among variants of the same scenario.
    }
    \label{fig:reward_ablation}
\end{figure*}

\stitle{\ours\ learns faster and keeps improving where plain RL plateaus.}
\emph{Timing}: in Figure~\ref{fig:reward_ablation}, both runs rise together for about 30 steps; then \textsc{Vanilla RL} peaks near step 60 and slips back, while \ours\ keeps climbing, matches RL's peak by mid-training, and ends about ten points ahead. \emph{Theory}: by Lemma~\ref{lem:credit}, groups with identical binary verifier rewards have zero group-relative reward advantages, so diagnosis weighting gives them no reward-driven GRPO policy-gradient signal; KL and other auxiliary terms may still contribute gradients, while \ours's interface can change later rollouts of those prompts. \emph{Diagnosis}: in a separate implementation-variant run (Appendix~\ref{app:failure_dynamics}), model-error label occurrences fall by over a third, led by response and state mismatches, while spurious-tool-call label counts stay approximately flat. \emph{Cost}: in a separate timing comparison, the online-interface configuration has 7.1\% higher median step time than Vanilla RL (Appendix~\ref{app:overhead}). In short, the same sparse reward keeps paying off under \ours\ well after RL stalls.

\subsection{Case Study}
\label{sec:case_study}

\stitle{The \ours-trained agent recovers where the base model gives up.}
Figure~\ref{fig:case_main} follows the BFCL task of Section~\ref{sec:intro} to the end. The base model fails twice: it gives up at the generic route error, leaving a wrong execution response, and later stores a GBP amount as a USD budget, leaving a wrong final state. The \ours-trained model repairs both and turns reward 0 into 1: it looks up the airport codes and retries, then converts the currency before setting the budget. This behavior illustrates the class-level recovery cue (``resolve the stated prerequisite before retrying'', Appendix~\ref{app:prompt_templates}), now followed without the cue at evaluation. Four further cases are in Appendix~\ref{app:case_study}.

\begin{figure*}[!t]
\centering
\scriptsize
\begin{minipage}[t]{0.42\textwidth}\vspace{0pt}
\begin{tcolorbox}[colback=red!5,colframe=red!45,boxrule=0.8pt,arc=2pt,left=4pt,right=4pt,top=3pt,bottom=3pt,equal height group=casebox,valign=top]
\textbf{Base model} (reward 0)\\[2pt]
\textbf{User:} economy flight cost, Rivermist to Stonebrook, 2026-10-06.\\
\textbf{Assistant} $\to$ \texttt{get\_flight\_cost(}\\ \hphantom{\textbf{Assistant} $\to$ }\texttt{travel\_from="RVM", travel\_to="STB", travel\_date=2026-10-06)}\\
\textbf{Tool:} \textcolor{red!70!black}{\texttt{Error: No available route for the given airports.}}\\
\textbf{Assistant:} ``There is no direct flight route available.''\\[2pt]
\textbf{User:} it is 15400 GBP; my access token is \texttt{abc123token}.\\
\textbf{Assistant} $\to$ \texttt{set\_budget\_limit(15400.0)} $\Rightarrow$ \textbf{Tool:} \texttt{\{"budget\_limit": 15400.0\}} \textcolor{red!70!black}{(GBP stored as USD)}
\end{tcolorbox}
\end{minipage}\hfill
\begin{minipage}[t]{0.56\textwidth}\vspace{0pt}
\begin{tcolorbox}[colback=green!6,colframe=green!45!black,boxrule=0.8pt,arc=2pt,left=4pt,right=4pt,top=3pt,bottom=3pt,equal height group=casebox,valign=top]
\textbf{\ours-trained model} (reward 1)\\[2pt]
\textbf{User:} economy flight cost, Rivermist to Stonebrook, 2026-10-06.\\
\textbf{Assistant} $\to$ \texttt{get\_flight\_cost(travel\_from="RVM", travel\_to="STB", travel\_date=2026-10-06)}\\
\textbf{Tool:} \textcolor{red!70!black}{\texttt{Error: No available route for the given airports.}}\\
\textbf{Assistant} $\to$ \texttt{get\_nearest\_airport\_by\_city("Rivermist")}, \texttt{("Stonebrook")} $\Rightarrow$ \textbf{Tool:} \texttt{"RMS"}, \texttt{"SBK"}\\
\textbf{Assistant} $\to$ \texttt{get\_flight\_cost("RMS", "SBK", 2026-10-06, economy)} $\Rightarrow$ \textbf{Tool:} \texttt{[220.0]}\\[2pt]
\textbf{User:} it is 15400 GBP; my access token is \texttt{abc123token}.\\
\textbf{Assistant} $\to$ \texttt{compute\_exchange\_rate(GBP, USD, 15400)} $\Rightarrow$ \textbf{Tool:} \texttt{22000.0}\\
\textbf{Assistant} $\to$ \texttt{set\_budget\_limit(22000.0)} $\Rightarrow$ \textbf{Tool:} \texttt{\{"budget\_limit": 22000.0\}}
\end{tcolorbox}
\end{minipage}
\vspace{-0.4em}
\caption{\small
\textbf{Before and after on BFCL \texttt{multi\_turn\_miss\_param\_150} (Qwen2.5-7B).}
Same task, tools, and verifier; only the policy differs. \textbf{Assistant} $\to$ lines are tool calls and \textbf{Tool} entries their abbreviated responses (one user turn omitted); the \ours\ trajectory matches the expected outcome (Appendix~\ref{app:case_api_parameter_recovery}).
}
\label{fig:case_main}
\vspace{-1.0em}
\end{figure*}
\FloatBarrier

\section{Conclusion}
\label{sec:conclusion}

This paper argues that a self-improving tool-use agent needs a learning environment that improves with it. We address Agent--Learning Environment Misalignment (ALEM) with \ours: a shared verifier-grounded diagnosis coordinates adaptive training support and diagnosis-aware policy updates, leaving the benchmark untouched. Across three backbones, this design improves over matched Vanilla RL; the factorial ablation yields a positive interaction point estimate, and the same checkpoints transfer positively to the evaluated held-out domains. In short, a diagnosis is a signal for the teacher as much as for the learner.



\clearpage

\subsection*{AI use statement}

Generative AI assisted experimental planning and refinement, implementation
and debugging, analysis and interpretation of rollout records, literature
search, and manuscript drafting and editing during development and revision.
An AI assistant also produced the first-pass annotations for the 60-case
blinded diagnosis audit in Appendix~\ref{app:diagnosis_audit}, using dialogue,
calls, initial state, and execution evidence while blinded to SEAL's label.
We report this as AI-assisted agreement, not independent human annotation.
Checks performed for this revision include recomputation of diagnostic
summaries from stored records, source/configuration consistency checks,
and LaTeX compilation. The training-time diagnoser itself remains deterministic and
does not invoke an LLM judge. The authors are responsible for the paper's
claims, results, code, and AI-assisted content.

\subsection*{Ethics statement}

This work studies tool-use agents in public benchmark environments
(BFCL V3, BFCL V4, and $\tau^2$-bench), rather than deployment with real
customers. More capable agents can be misused, and improved benchmark
performance does not establish safe deployment. Training guidance excludes
explicit reference calls and hidden answer values, but the finite-template
analysis is not a guarantee against information leakage. Benchmark and
model use remains subject to the respective released licenses.

\subsection*{Reproducibility statement}

Appendices~\ref{app:benchmark_details}--\ref{app:experimental_details}
describe the scenario-level split, algorithm, qualified theoretical
properties, hyperparameters, and diagnostic rules.
Appendix~\ref{app:prompt_templates} supplies interface templates, and
Appendix~\ref{app:revision_analysis} describes the main comparisons
and auxiliary analyses. The main Qwen2.5-7B configuration is specified
in Appendix~\ref{app:run_correspondence}; the auxiliary diagnosis audit's
protocol is described in Appendix~\ref{app:diagnosis_audit}.
Code and split manifests are intended for release. The training
comparisons report one run per condition and do not estimate
variation across training seeds.

\clearpage
\bibliographystyle{unsrtnat}
\bibliography{references}

\clearpage
\beginappendix

\section{Benchmark and Evaluation Details}
\label{app:benchmark_details}

\stitle{BFCL V3.}
Our in-distribution evaluation uses the multi-turn subset of the Berkeley Function-Calling Leaderboard (BFCL) V3. The benchmark evaluates whether an agent can correctly use executable tools over multi-turn dialogue contexts. Each task contains a user instruction sequence, a set of available tools, and an executable evaluation protocol. The subset contains 800 examples across four categories: \textit{Base}, \textit{Missing Functions}, \textit{Missing Parameters}, and \textit{Long Context}. \textit{Base} evaluates standard multi-turn function calling with complete tool information. \textit{Missing Functions} tests whether the agent can recognize when required tools are unavailable or when no valid tool applies. \textit{Missing Parameters} evaluates whether the agent can handle underspecified arguments rather than hallucinating missing values. \textit{Long Context} stresses state tracking and tool-use consistency over extended dialogue histories.

The 800 examples comprise 200 scenario IDs across the four categories,
with 200 examples per category. Following the 400-training / 400-evaluation
setup used by Environment Tuning~\citep{lu2026dont}, we construct our
scenario-disjoint split by shuffling the IDs with Python seed 42 and
assigning 100 to training and 100 to evaluation across all four categories.
This gives 400 examples in each partition, with 100 per category. The final
evaluation split is not used for policy training or interface adaptation.
The four subsets are benchmark categories, not component ablations. All
compared methods use the same split. The task-ID manifests and split checks
specify the data independently of the random seed used for policy training.

\stitle{BFCL V4.}
For out-of-distribution evaluation, we use the Web Search and Memory tracks from BFCL V4. These tracks differ from BFCL V3 in tool schema structure, task distribution, and interaction pattern. The Web Search track evaluates whether an agent can retrieve and synthesize information using search-style tools. The Memory track evaluates whether an agent can store, retrieve, and update memory-like information across interactions. Since SEAL is trained only on the BFCL V3 training split, BFCL V4 serves as a held-out benchmark for testing cross-distribution transfer.

\stitle{$\tau^2$-bench.}
We also evaluate on $\tau^2$-bench~\citep{barres2025tau2} across the Retail, Airline, and Telecom domains. These tasks require multi-turn tool use in simulated customer-service environments, where the agent must follow domain policies, interact with tools, and maintain task state across turns. Compared with BFCL, $\tau^2$-bench introduces different tool semantics, domain constraints, and user-agent interaction patterns, making it a complementary testbed for OOD generalization.

\stitle{Evaluation protocol and metric.}
All evaluations are performed using the original benchmark environments and verifiers. SEAL-specific diagnostic feedback, interface augmentation, capability-specific hints, and recovery-oriented training messages are disabled during evaluation. Thus, all reported results measure the policy under the original test-time protocol. We report success rate as the primary metric:
\begin{equation}
\mathrm{SR}
=
\frac{1}{|\mathcal{Q}|}
\sum_{q\in\mathcal{Q}}
\mathbf{1}
\left[
\mathcal{V}(\tau_q,q,\mathcal{E})=1
\right],
\end{equation}
where $\mathcal{Q}$ is the evaluation set and $\tau_q$ is the trajectory generated for instruction $q$. Category-level scores are computed on the corresponding subset, and average scores are computed as the arithmetic mean over reported categories.

\section{Backbones and Reference Systems}
\label{app:models}

\stitle{Trained backbones.}
Qwen2.5-3B-Instruct and Qwen2.5-7B-Instruct \citep{yang2024qwen25} are
general-purpose instruction-tuned models from the Qwen2.5 family and serve
as the two general backbones. ToolACE-2-Llama-3.1-8B \citep{liu2025toolace}
is a Llama-3.1-8B model post-trained on ToolACE function-calling data and
serves as the tool-specialized backbone. All three are trained with
\textsc{Vanilla RL} and with \ours\ under the protocol of
Section~\ref{sec:exp_settings}.

\stitle{Method-level baseline.}
Environment Tuning \citep{lu2026dont} improves multi-turn tool use by
changing the training environment rather than supplying expert
trajectories: a four-stage curriculum (format and tool-call rewards, then
the Base split, then the missing-function, missing-parameter, and
long-context splits, then all scaffolding removed), actionable error
augmentation that replaces uninformative execution errors with pedagogical
hints, and a per-turn progress reward averaged over the episode, trained
with a GRPO variant. We run it on the same backbones and 400-example split
as \ours. It differs from \ours\ in changing the reward and using
hand-written per-domain hints, whereas \ours\ keeps the binary verifier
reward and derives its cues from diagnosed failure classes.

\stitle{Reference systems in Table~\ref{tab:bfcl_v3_results}.}
These rows are reported only to bound the attainable range on BFCL V3; they
differ from our backbones in scale, training data, and recipe and are not
budget-matched. Closed-source: Gemini-3-Pro-Preview
\citep{google2025gemini3propreview} and Gemini-2.5-Flash
\citep{comanici2025gemini25} from Google, Claude-Sonnet-4.5
\citep{anthropic2025claudesonnet45} from Anthropic, and GPT-4o-2024-11-20
\citep{hurst2024gpt4o} from OpenAI. Open-source: GLM-4.6 (355B)
\citep{glm46modelcard2025}, Qwen3-32B \citep{yang2025qwen3},
Qwen2.5-14B-Instruct \citep{yang2024qwen25}, the larger sibling of our Qwen
backbones, and xLAM-2-3b-fc-r \citep{xlam2modelcard2025}, a 3B
function-calling model trained on large-scale synthetic multi-turn
tool-use data, which is why it sits far above our 400-example 3B rows.

\stitle{Reference systems in Table~\ref{tab:ood_results}.}
BitAgent-8B \citep{bitagent8b2025} and watt-tool-8B \citep{watttool8b2024}
are open 8B tool-calling models; ToolACE-2-Llama-3.1-8B
\citep{liu2025toolace} is our tool-specialized backbone evaluated without
training; Qwen3-30B-A3B-Instruct \citep{yang2025qwen3} is a mixture-of-experts
model with about 3B active parameters. None of these rows is budget-matched
to our trained models.

\section{SEAL Training Algorithm}
\label{app:algorithm}

Algorithm~\ref{alg:seal} specifies the full SEAL training procedure.
SEAL alternates between policy rollout, verifier-grounded diagnosis,
diagnosis-guided interface evolution, and policy optimization. The
interface wrapper is used only during training and is removed during final
evaluation. The main Qwen2.5-7B SEAL checkpoint uses both online recovery
selection and diagnosis-aware policy weighting. Its correspondence to
the reported results is specified in Appendix~\ref{app:run_correspondence}.

\begin{algorithm}[t]
\caption{SEAL: Synergistic Co-Evolution of Agents and
Learning Environments}
\label{alg:seal}
\begin{algorithmic}[1]
\Require Initial policy $\pi_{\theta_0}$, reference policy $\pi_{\mathrm{ref}}$,
training prompts $\mathcal{D}$, base environment $\mathcal{E}$, verifier $\mathcal{V}$,
number of rounds $R$, rollouts per prompt $G$
\Require Initial interface wrapper $\Omega_0$, base utilities $\rho_0(z)$,
clipping range $[w_{\min}, w_{\max}]$
\For{$r = 0,1,\ldots,R-1$}
    \State Sample prompt batch $\mathcal{X}_r \subset \mathcal{D}$
    \State Construct training environment $\mathcal{E}_{\Omega_r} \gets \Omega_r(\mathcal{E})$
    \State Initialize rollout buffer $\mathcal{R}_r \gets \emptyset$
    \For{each prompt $x \in \mathcal{X}_r$}
        \State Sample $G$ trajectories from $\pi_{\theta_r}$ in $\mathcal{E}_{\Omega_r}$
        \For{each trajectory $\tau_j$}
            \State Compute verifier reward $r_j \gets \mathcal{V}(\tau_j,x,\mathcal{E})$
            \State Diagnose trajectory $Z(\tau_j) \gets \Psi(\tau_j,\mathcal{E},\mathcal{V})$
            \State Store $(x,\tau_j,r_j,Z(\tau_j))$ in $\mathcal{R}_r$
        \EndFor
    \EndFor
    \State Aggregate failure profile $C_r \gets \operatorname{Agg}(\{Z(\tau_j): \tau_j \in \mathcal{R}_r\})$
    \State Update $\rho_r$ once using Eqs.~\eqref{eq:utility_target}--\eqref{eq:utility_update}
    \For{each prompt group in $\mathcal{R}_r$}
        \State Compute group-relative advantages $A_j$ from verifier rewards $\{r_j\}_{j=1}^{G}$
        \State Compute weights $w_j \gets \operatorname{clip}(\sum_{z\in\mathcal{Z}} p_{\tau_j}(z)\rho_r(z), w_{\min}, w_{\max})$
    \EndFor
    \State Set weighted advantages $\widetilde{A}_j \gets w_j A_j$ for every trajectory
    \State Update policy $\pi_{\theta_r}\rightarrow\pi_{\theta_{r+1}}$ using GRPO and KL regularization
    \State Update interface $\Omega_{r+1}\gets U(\Omega_r,C_r)$
\EndFor
\State \textbf{return} trained policy $\pi_{\theta_R}$
\end{algorithmic}
\end{algorithm}

One round is one optimizer iteration with 32 prompts and eight trajectories
per prompt. Diagnoses determine the current trajectories' weights and the
next round's interface. There is no retroactive change to sampled actions
or tool outputs.

At evaluation time, SEAL discards the training-time interface wrapper and
evaluates $\pi_{\theta_R}$ in the original environment $\mathcal{E}$ with
the unchanged verifier $\mathcal{V}$.

\FloatBarrier
\section{Theoretical Analysis}
\label{app:theory}

This appendix states what the update rules of Section~\ref{sec:method} do
and do not guarantee. Every statement concerns the rules as specified, at
the population or single-batch level; none is a finite-sample or
convergence guarantee for the full training loop. Notation follows
Section~\ref{sec:method}: a prompt yields a group of $G$ rollouts with
binary verifier rewards $r_j\in\{0,1\}$ and group-normalized advantages
$A_j$; each trajectory has a diagnostic profile $p_j$ over labels
$\mathcal{Z}$; utilities $\rho_t$ follow
Eqs.~\eqref{eq:utility_target}--\eqref{eq:utility_update}; weights satisfy
$w_j\in[w_{\min},w_{\max}]=[0.5,2.0]$; and the augmented observation is
$\tilde o_i=\Omega_t(o_i,\mathcal{F},H_t,C_{t-1})$.

\subsection{Credit Under a Binary Reward}
\label{app:theory-credit}

\begin{lemma}[Binary GRPO credit is failure-blind]
\label{lem:credit}
Consider a group of $G$ rollouts of one prompt with $k$ successes,
$0<k<G$, and advantages $A_j=(r_j-\mu)/\sigma$ with $\mu=k/G$ and
$\sigma=\sqrt{\mu(1-\mu)}$. Then every successful trajectory receives
$A^{+}=\sqrt{(G-k)/k}$ and every failed trajectory receives
$A^{-}=-\sqrt{k/(G-k)}$. Hence the coefficient of
$\nabla_\theta\log\pi_\theta(\tau_j)$ in the unclipped reward-surrogate
gradient at the rollout policy is the same for all failed trajectories
of the group, whatever their failure modes. If $k\in\{0,G\}$, this reward
term contributes zero; a separate KL term need not vanish.
\end{lemma}
\begin{proof}
For $r_j=1$,
$A_j=(1-\mu)/\sqrt{\mu(1-\mu)}=\sqrt{(1-\mu)/\mu}=\sqrt{(G-k)/k}$; for
$r_j=0$, $A_j=-\mu/\sqrt{\mu(1-\mu)}=-\sqrt{k/(G-k)}$. If $k\in\{0,G\}$
then $\sigma=0$ and GRPO assigns zero advantage to the group. Using the
unbiased sample standard deviation instead of $\sigma$ rescales every
advantage of the group by the common factor $\sqrt{(G-1)/G}$, preserving
equality within each reward class, but rescaling the displayed values.
\end{proof}

The lemma formalizes challenge \ctwo\ of the introduction: under a
terminal binary reward, the scalar group-normalized advantage cannot
distinguish an invalid argument from a wrong final answer. An additional
weight $w_j$ can distinguish their diagnostic profiles. This statement
does not identify the full token-level, clipped, KL-regularized update
with a trajectory score-function gradient.

\begin{remark}[Division of labor between the two sides]
\label{rem:division}
If all binary verifier rewards in a prompt group match, every group-relative
reward advantage is zero. Multiplying by $w_j$ therefore leaves the group's
reward-driven GRPO policy-gradient contribution at zero; KL and other
auxiliary regularizers may still contribute gradients. Reweighting changes
the reward signal only in mixed-success groups. The interface $\Omega_t$
instead changes observations for subsequent sampling, potentially also
for currently unsolved prompts. Updates from other prompts and KL
regularization can also change their future rollouts. These are distinct
interventions, not a proof that either improves learning or that their
effects are independent.
\end{remark}

\subsection{Utility Dynamics}
\label{app:theory-utility}

\begin{proposition}[Utility dynamics]
\label{prop:utility}
Let $\rho_0(z)\in[0.5,2.0]$ for all $z\in\mathcal{Z}$ and let $\rho_t^*$
and $\rho_t$ follow Eqs.~\eqref{eq:utility_target}
and~\eqref{eq:utility_update}. For every non-pass label $z$ and round $t\ge1$:
\begin{enumerate}[label=(\alph*),topsep=1pt,itemsep=1pt,leftmargin=2em]
\item \emph{Invariance.} $\rho_t(z)\in[0.5,2.0]$.
\item \emph{Bounded drift.} $|\rho_t(z)-\rho_{t-1}(z)|\le0.25\,(2.0-0.5)=0.375$.
\item \emph{Bounded, monotone target.}
$\rho_0(z)\le\rho_t^*(z)\le\min\{2\rho_0(z),\,2.0\}$, and $\rho_t^*(z)$ is
nondecreasing in the label's share $p_t(z)$, in its co-failure rate
$1-\bar r_t(z)$, and in its support $n_t(z)$.
\item \emph{Geometric tracking.} If $\rho_s^*(z)=\rho^*(z)$ for all
$s\in(t_0,t]$, then
$|\rho_t(z)-\rho^*(z)|\le0.75^{\,t-t_0}\,|\rho_{t_0}(z)-\rho^*(z)|$; if
instead the target moves by at most $\delta$ per round, then
$\limsup_{t}|\rho_t(z)-\rho_t^*(z)|\le3\delta$.
\item \emph{Self-limiting emphasis.} If $p_t(z)\to0$ or
$\bar r_t(z)\to1$, then $\rho_t^*(z)\to\rho_0(z)$ and
$\rho_t(z)\to\rho_0(z)$.
\end{enumerate}
\end{proposition}
\begin{proof}
(a) By induction: $\rho_t^*(z)\in[0.5,2.0]$ by the clip, and $\rho_t(z)$
is a convex combination of $\rho_{t-1}(z)$ and $\rho_t^*(z)$.
(b) $\rho_t-\rho_{t-1}=0.25\,(\rho_t^*-\rho_{t-1})$, and both terms lie in
$[0.5,2.0]$.
(c) Each of $p_t(z)$, $1-\bar r_t(z)$, and $c_t(z)$ lies in $[0,1]$, so the
bracket in Eq.~\eqref{eq:utility_target} lies in $[\rho_0(z),2\rho_0(z)]$
and is nondecreasing in each factor; clipping preserves monotonicity, and
the lower clip is inactive because $\rho_0(z)\ge0.5$.
(d) Let $e_t=\rho_t(z)-\rho_t^*(z)$. Eq.~\eqref{eq:utility_update} gives
$e_t=0.75\,\big(e_{t-1}+\rho_{t-1}^*(z)-\rho_t^*(z)\big)$. With a constant
target the second term vanishes and $e_t=0.75\,e_{t-1}$; with
$|\rho_{t-1}^*(z)-\rho_t^*(z)|\le\delta$,
$|e_t|\le0.75\,|e_{t-1}|+0.75\,\delta$, whose fixed point is $3\delta$.
(e) The bracket tends to $\rho_0(z)$, so $\rho_t^*(z)\to\rho_0(z)$, and by
(d) the moving average converges to the same limit.
\end{proof}

In words, the emphasis on a failure type rises while that type is frequent
and unproductive, can at most double its base value, moves by at most
$0.375$ per round, closes 90\% of the gap to a stationary target within
eight rounds ($0.75^{8}\approx0.10$), and returns to the base value once
the type disappears from the on-policy rollouts. The last property is what
reduces persistent emphasis after a label disappears; it does not prove
that the method fixes a failure. The variant observations in
Table~\ref{tab:failure_windows} are not a validation of these dynamics.

\subsection{Reweighting as a Bounded Tilt of the Rollout Distribution}
\label{app:theory-tilt}

\begin{proposition}[Bounded tilt]
\label{prop:tilt}
Let $w:\tau\mapsto[w_{\min},w_{\max}]$ with $0<w_{\min}\le w_{\max}$ and
let $c(\tau)$ be any coefficient function; both are treated as constants
with respect to $\theta$, as the GRPO advantage is. Define
$Z_w=\mathbb{E}_{\tau\sim\pi_\theta}[w(\tau)]$ and the tilted distribution
$q_w(\tau)=w(\tau)\,\pi_\theta(\tau)/Z_w$. Then
\begin{equation}
\mathbb{E}_{\tau\sim\pi_\theta}\big[w(\tau)\,c(\tau)\,\nabla_\theta\log\pi_\theta(\tau)\big]
= Z_w\;\mathbb{E}_{\tau\sim q_w}\big[c(\tau)\,\nabla_\theta\log\pi_\theta(\tau)\big],
\label{eq:tilt}
\end{equation}
and (i) $q_w$ and $\pi_\theta$ have the same support; (ii)
$w_{\min}/w_{\max}\le q_w(\tau)/\pi_\theta(\tau)\le w_{\max}/w_{\min}$ for
every $\tau$; (iii) $\mathrm{KL}(q_w\,\|\,\pi_\theta)\le\log(w_{\max}/w_{\min})$
and $\mathrm{KL}(\pi_\theta\,\|\,q_w)\le\log(w_{\max}/w_{\min})$.
\end{proposition}
\begin{proof}
Eq.~\eqref{eq:tilt} follows from $q_w(\tau)=w(\tau)\pi_\theta(\tau)/Z_w$
by substitution. (i) $w>0$. (ii) $q_w/\pi_\theta=w/Z_w$ and
$w_{\min}\le Z_w\le w_{\max}$. (iii)
$\mathrm{KL}(q_w\|\pi_\theta)=\mathbb{E}_{q_w}[\log(w/Z_w)]\le\log(w_{\max}/w_{\min})$
by (ii), and likewise
$\mathrm{KL}(\pi_\theta\|q_w)=\mathbb{E}_{\pi_\theta}[\log(Z_w/w)]\le\log(w_{\max}/w_{\min})$.
\end{proof}

With \ours's range $[0.5,2.0]$, the density ratio lies in $[1/4,4]$ and
both divergences are at most $\log 4\approx1.39$ nats. The
weighted score-function term can therefore be represented under a
distribution that emphasizes high-utility trajectories, with prefactor
$Z_w\in[0.5,2.0]$. This algebra neither changes the sampling procedure
nor equates the full clipped, KL-regularized GRPO objective to an
unweighted objective. Positive weights preserve each advantage's sign.
The corresponding finite-batch identity is
$\sum_j w_jA_j\nabla_\theta\log\pi_\theta(\tau_j)=W\sum_j(w_j/W)\,A_j\nabla_\theta\log\pi_\theta(\tau_j)$
with $W=\sum_jw_j\in[Nw_{\min},Nw_{\max}]$. The prefactor is bounded,
but the ratio of weighted to unweighted aggregate gradient norms is not:
unequal weights can change cancellation between trajectory gradients.

\begin{corollary}[The utility-average clip never binds]
\label{cor:convex}
Since $p_j$ is a probability distribution over labels and
$\rho_t(z)\in[0.5,2.0]$ for all $z$ (Proposition~\ref{prop:utility}(a)),
$\sum_z p_j(z)\rho_t(z)$ is a convex combination of utilities and lies in
$[\min_z\rho_t(z),\max_z\rho_t(z)]\subseteq[0.5,2.0]$. The clip in the
definition of $w_j$ is therefore inactive. The direct multiplier remains
in $[0.5,2.0]$, so the density-ratio bound of 4 in
Proposition~\ref{prop:tilt}(ii) applies.
\end{corollary}

\subsection{Interface Evolution: Policy Preservation and a Finite Channel}
\label{app:theory-interface}

\begin{proposition}[Interface evolution]
\label{prop:interface}
Let
$\tilde o_i=o_i\oplus\phi_{\mathrm{schema}}(\mathcal{F})\oplus\phi_{\mathrm{err}}(o_i,H_t)\oplus\phi_{\mathrm{cap}}(C_{t-1})$,
where $\phi_{\mathrm{schema}}$ is a deterministic function of the tool
schema, and $\phi_{\mathrm{err}}$ and $\phi_{\mathrm{cap}}$ instantiate
fixed templates chosen, as a function of $(H_t,C_{t-1})$, from a finite set
$\mathcal{L}$ of complete template configurations, including cue presence,
ordering, and placement within one round; the transition dynamics, the tool
outputs, and the reward $r(\tau)=\mathcal{V}(\tau,q,\mathcal{E})$ are
unchanged. Then:
\begin{enumerate}[label=(\roman*),topsep=1pt,itemsep=1pt,leftmargin=2.2em]
\item \emph{Policy preservation.} For every unrestricted policy $\pi$ over original
observations, the policy $\tilde\pi(a\mid\tilde o_{\le i})=\pi(a\mid o_{\le i})$
over augmented observations induces the same trajectory distribution;
hence $\sup_{\tilde\pi}J(\tilde\pi)\ge\sup_{\pi}J(\pi)$, and the set of
verifier-accepted trajectories is identical with and without the interface.
\item \emph{Finite-channel bound.} Let $R_q$ be the reference of task $q$ (its
reference call sequence, hidden parameter values, and final answer), and
let $\ell\in\mathcal{L}$ be the template configuration selected in round
$t$. Conditionally on the agent's own observation history $o_{\le i}$ and
the schema $\mathcal{F}$, the augmented observations depend on $R_q$
(through the verifier-derived labels that shape $H_t$ and $C_{t-1}$) only via
$\ell$, so
$I(R_q;\tilde o_{\le i}\mid o_{\le i},\mathcal{F})\le H(\ell\mid o_{\le i},\mathcal{F})\le\log_2|\mathcal{L}|$
bits per round.
\end{enumerate}
\end{proposition}
\begin{proof}
(i) $o_i$ is the first component of $\tilde o_i$, so $\tilde\pi$ is well
defined and reproduces the action distribution of $\pi$ at every turn; the
transition kernel and the verifier do not read $\tilde o_i$, so the induced
trajectory distribution and its return coincide, which gives the inequality
after taking suprema. Acceptance depends only on $(\tau,q,\mathcal{E})$.
(ii) Given $(o_{\le i},\mathcal{F})$, $\tilde o_{\le i}$ is a deterministic
function of $\ell$; the data-processing inequality gives
$I(R_q;\tilde o_{\le i}\mid o_{\le i},\mathcal{F})\le I(R_q;\ell\mid o_{\le i},\mathcal{F})\le H(\ell\mid o_{\le i},\mathcal{F})\le\log_2|\mathcal{L}|$.
\end{proof}

Part (i) concerns unrestricted policies and append-only observations;
it is not a guarantee for a fixed-capacity model or a context window that
truncates original observations. Part (ii) is a conditional, single-round
bound. The cardinality must count complete configurations, not just label
names or individual strings. Even one bit can reveal an answer among two
candidates, and information may accumulate across rounds; the bound is
not a no-leakage guarantee. A transcript of $T$ configurations has at most
$T\log_2|\mathcal{L}|$ bits of entropy under the same finite-configuration
assumption. Neither part guarantees performance after removing training
cues. The removed-at-evaluation results are empirical observations, not
a consequence of these propositions.

\section{Experimental Details}
\label{app:experimental_details}

This section provides implementation details for reproducing the training
runs and the controlled Vanilla RL comparison. Within each backbone, the
binary-reward comparison uses the same training split, rollout budget,
decoding configuration, and evaluation verifier. Environment Tuning
follows its released curriculum, reward design, and training recipe and
serves as a method-level baseline. Vanilla RL provides the matched
binary-reward control under the same GRPO backbone and rollout protocol.

\subsection{Hyperparameter Configuration}

Table~\ref{tab:hyperparameters} summarizes the main training configuration
used in our BFCL experiments. We use Qwen2.5-7B-Instruct as the default
backbone for the main BFCL experiments and apply the same GRPO
configuration to Vanilla RL and SEAL. The only algorithmic differences are
whether verifier-grounded diagnosis, diagnosis-guided interface evolution,
closed-loop failure-profile updates, and diagnostic advantage weighting
are enabled.

\begin{table*}[t]
    \centering
    \caption{\small
    \textbf{Main hyperparameter configuration for BFCL experiments.}
    Unless otherwise specified, Vanilla RL and SEAL use the same optimization,
    rollout, and evaluation settings.
    }
    \label{tab:hyperparameters}
    \small
    \setlength{\tabcolsep}{6pt}
    \renewcommand{\arraystretch}{1.12}
    \begin{tabular}{llc}
        \toprule
        \textbf{Category} & \textbf{Hyperparameter} & \textbf{Value} \\
        \midrule

        \multirow{6}{*}{Policy optimization}
        & Base model & Qwen2.5-7B-Instruct \\
        & Training method & GRPO \\
        & Actor learning rate & $1\times10^{-6}$ \\
        & Train batch size & 32 \\
        & PPO mini-batch size & 8 \\
        & PPO micro-batch size / GPU & 1 \\
        \midrule

        \multirow{7}{*}{Rollout}
        & Inference engine & vLLM (async) \\
        & Rollouts per prompt & 8 \\
        & Training temperature & 1.0 \\
        & Validation temperature & 0.0 \\
        & Max prompt length & 8192 \\
        & Max response length & 4096 \\
        & Max model length & 16384 \\
        \midrule

        \multirow{5}{*}{Multi-turn setting}
        & Number of training samples & 400 \\
        & Multi-turn rollout & Enabled \\
        & Max interaction steps & 20 \\
        & Tool format & \texttt{llama3\_json} \\
        & BFCL tool prompt mode & \texttt{bfcl\_qwen\_fc} \\
        \midrule

        \multirow{4}{*}{Regularization}
        & Reward KL penalty & Disabled \\
        & Actor KL loss coefficient & 0.001 \\
        & Entropy coefficient & 0.0 \\
        \midrule

        \multirow{4}{*}{Hardware}
        & GPUs & 4 \\
        & Tensor parallel size & 1 \\
        & GPU memory utilization & 0.6 \\
        & Max environment workers & 32 \\

        \bottomrule
    \end{tabular}
\end{table*}

For comparisons with different backbones, we keep the same data split,
rollout budget, verifier, decoding configuration, and evaluation protocol
whenever possible. When hardware constraints require changes to
micro-batch size or gradient accumulation, we keep the effective batch size
fixed.

\subsection{Diagnosis Label Priority}
\label{app:diagnosis_priority}

Table~\ref{tab:diagnosis_labels} in the main text specifies the operational
decision sequence together with the fixed base utilities $\rho_0$. These
seven outcomes are the labels used for training; the generic behavioral
terms in Figure~\ref{fig:seal_training_loop} are not a second taxonomy. An
exception while executing a reference program or running a checker produces
\texttt{gt\_error} or \texttt{checker\_error}; these bookkeeping labels are
excluded from the learning utilities. Required tool use and acceptance are
defined by the original executable verifier, and reference information
stays inside the checker.

This ordering separates absence of a required call from state and response
failures. It does not assert that every invalid argument has a unique
diagnostic label, or that every downstream mismatch is a new causal error.
Model-side and reference-side execution use separate state instances.

The trajectory weight averages valid diagnostic utilities and directly
scales the normalized group-relative advantage, without an additional
outcome-calibration stage or subsequent advantage renormalization. The
same trajectory-level multiplier applies to positive and negative
advantages and to all response tokens. Base utilities, EMA coefficient
0.25, and support constant 32 are fixed across the matched experiments.
For example, two correct abstentions and one state mismatch give
$w=(1.8+1.8+1.0)/3\approx1.53$ under the base utilities. This
trajectory-level form preserves the underlying GRPO objective's
token-level loss structure.

\subsection{Controlled Comparison Protocol}

The comparison between Vanilla RL and SEAL is controlled along the
following dimensions. Both methods use the same initial checkpoint, the
same set of 400 BFCL V3 training samples, the same number of rollouts per
prompt, the same optimizer, and the same decoding configuration. Both
methods are evaluated using the original BFCL V3 multi-turn environment
and the official verifier.
These matches concern examples, rollouts, and optimization settings;
wall-clock time and prompt-token exposure can differ because SEAL adds
training-time interface content.

Vanilla RL optimizes the standard GRPO objective using only verifier
rewards. SEAL uses the same verifier rewards and the same GRPO backbone,
but additionally uses verifier-grounded diagnoses from training rollouts
to construct training-time interface updates and diagnostic utility
weights. No policy receives ground-truth tool trajectories, hidden
parameter values, reference intermediate actions, or final answers as
training observations; the executable verifier internally uses benchmark
references to compute outcomes.

The evolved SEAL interface is used only during training. During
evaluation, the wrapper is removed, and the model interacts with the
original benchmark interface. Thus, SEAL and Vanilla RL are evaluated
under identical test-time conditions. The comparison therefore measures
whether the policy has internalized better tool-use behavior, rather than
whether it has access to additional test-time information.

\subsection{Logging and Reproducibility}

For each reported training run, we record the random seed, base checkpoint,
code version, training configuration and overrides, rollout budget,
decoding parameters, round-level diagnostic profiles, interface states,
validation scores, and final per-example evaluation outputs. These records
provide run-level correspondence between the training configuration,
checkpoint, and reported evaluation result.

\subsection{Run and Configuration Correspondence}
\label{app:run_correspondence}

\stitle{Main Qwen2.5-7B checkpoint.}
The Qwen2.5-7B SEAL result in Table~\ref{tab:bfcl_v3_results} is the
full-online configuration specified in Algorithm~\ref{alg:seal}. During
training, it uses fixed required-argument and enum schema affordances,
online diagnosis-conditioned recovery selection, and diagnosis-aware
advantage reweighting.

At round $t$, trajectories are collected under the current interface
$\Omega_t$. Their verifier-grounded diagnoses determine the advantage
weights for the current policy update and the recovery guidance selected
for $\Omega_{t+1}$. Schema affordances remain fixed throughout training,
while recovery guidance evolves from preceding on-policy diagnoses.

The resulting checkpoint scores 161/400 (40.25\%) on the BFCL V3
evaluation split, with Base, Missing Functions, Missing Parameters, and
Long Context accuracies of 58.00\%, 36.00\%, 34.00\%, and 33.00\%,
respectively. The Qwen2.5-7B SEAL OOD results in Table~\ref{tab:ood_results}
evaluate this same checkpoint without additional training or checkpoint
selection. All SEAL-specific interface additions are removed during
evaluation, and the checkpoint is evaluated with the original BFCL
environment and verifier. Factorial control configurations are specified in
Appendix~\ref{app:matched_controls}.

\section{Prompt and Interface Templates}
\label{app:prompt_templates}

This section reports the prompt and training-time interface templates used
in our BFCL implementation. The base tool-use prompt is shared by Vanilla
RL and SEAL. SEAL modifies only the training-time environment interface,
for example by annotating tool schemas, enriching failed tool feedback when
enabled, or injecting short behavioral guidelines derived from recent
diagnostic profiles. All SEAL-specific interface augmentations are disabled
during evaluation.

\subsection{Base BFCL Tool-Use Prompt}

For Qwen-style BFCL experiments, the environment serializes the available
tools into the system prompt using the following template. This base prompt
is used by both Vanilla RL and SEAL.

\begin{tcolorbox}[
        breakable,
        title=BFCL Qwen Tool Prompt Template,
    ]
    \begin{lstlisting}[breaklines=true, breakatwhitespace=true, basicstyle=\small\ttfamily, gobble=4, breakautoindent=true]
    # Tools

    You may call one or more functions to assist with the user query.

    You are provided with function signatures within <tools></tools> XML tags:
    <tools>
    $tool_schema_json_1
    $tool_schema_json_2
    ...
    </tools>

    For each function call, return a json object with function name and arguments within <tool_call></tool_call> XML tags:
    <tool_call>
    {"name": <function-name>, "arguments": <args-json-object>}
    </tool_call>
    \end{lstlisting}
\end{tcolorbox}

The base prompt defines the tool-call format and exposes the available tool
schemas. It does not contain SEAL-specific diagnostic labels, recovery
hints, or capability-dependent instructions.

\subsection{Schema Observation Annotation}

In our active BFCL configuration, SEAL uses a lightweight observation
annotation mechanism. When \texttt{observation\_lite} is enabled, SEAL does
not change tool APIs, tool signatures, tool outputs, or verifier behavior.
Instead, it augments the exposed schema descriptions with lightweight
annotations that make existing constraints more explicit.

\begin{tcolorbox}[
        breakable,
        title=Schema Observation Annotation Template,
    ]
    \begin{lstlisting}[breaklines=true, breakatwhitespace=true, basicstyle=\small\ttfamily, gobble=4, breakautoindent=true]
    $argument_description [required]

    $argument_description Allowed values: $enum_value_1, $enum_value_2, ...
    \end{lstlisting}
\end{tcolorbox}

The environment also appends the following training-time instruction:

\begin{tcolorbox}[
        breakable,
        title=Observation-Lite Instruction,
    ]
    \begin{lstlisting}[breaklines=true, breakatwhitespace=true, basicstyle=\small\ttfamily, gobble=4, breakautoindent=true]
    Observation note: parameters marked [required] must be provided exactly when calling a tool; do not invent tools outside the current schema.
    \end{lstlisting}
\end{tcolorbox}

This annotation corresponds to the schema-affordance component
$\phi_{\mathrm{schema}}$. It exposes only public schema information, such
as required arguments and enum constraints, and does not reveal the correct
tool choice, hidden parameter values, reference tool trajectory, or final
answer for any instance.

\subsection{Tool Feedback Evolution Template}

When \texttt{tool\_feedback\_evolution} is enabled, SEAL enriches only
failed interactions. This component corresponds to the recovery-feedback
component $\phi_{\mathrm{err}}$. It converts low-level parse or execution
errors into structured recovery hints, while preserving the original tool
API and verifier.

For parser failures, the environment may append format-specific hints such
as:

\begin{tcolorbox}[
        breakable,
        title=Parse Error Hint Template,
    ]
    \begin{lstlisting}[breaklines=true, breakatwhitespace=true, basicstyle=\small\ttfamily, gobble=4, breakautoindent=true]
    $base_parse_error

    [Hint] Each tool call MUST be a JSON object inside <tool_call>...</tool_call> tags with both "name" and "arguments" keys, e.g. <tool_call>{"name": "<tool>", "arguments": {"<arg>": <value>}}</tool_call>.
    [Hint] Available tools: $tool_name_1, $tool_name_2, ...
    [Hint] If no available tool can serve the user's request, reply in plain text instead of producing a malformed tool_call.
    \end{lstlisting}
\end{tcolorbox}

For execution errors, the environment may append tool-specific but
non-solution-revealing hints such as:

\begin{tcolorbox}[
        breakable,
        title=Tool Error Hint Template,
    ]
    \begin{lstlisting}[breaklines=true, breakatwhitespace=true, basicstyle=\small\ttfamily, gobble=4, breakautoindent=true]
    [ToolHint] Required argument(s) $args for `$tool_name` were missing. Re-emit the tool_call with these keys present in `arguments`.

    [ToolHint] `$tool_name` does not accept argument `$arg`. Drop it and only use arguments listed in the tool schema.

    [ToolHint] `$tool_name` rejected an argument type. Re-check argument types in the tool schema (string vs int vs list).
    \end{lstlisting}
\end{tcolorbox}

These hints identify the class of tool-use error and the violated
interface constraint. They do not provide the next correct tool call,
instance-specific argument values, or the ground-truth solution trajectory.

\subsection{Diagnostic Evolution Guideline Template}

When \texttt{diagnostic\_evolution} is enabled, SEAL reads recent failure
statistics from the capability profile and injects short behavioral
guidelines into the system prompt. This component corresponds to
$\phi_{\mathrm{cap}}$: the interface is adjusted according to recurring
failure modes of the current policy rather than individual benchmark
instances.

\begin{tcolorbox}[
        breakable,
        title=Diagnostic-Driven System Guideline Template,
    ]
    \begin{lstlisting}[breaklines=true, breakatwhitespace=true, basicstyle=\small\ttfamily, gobble=4, breakautoindent=true]
    # Behavioral guidelines (env-evolved from recent failures):
    - [Cautious-call] If no available function clearly matches the user's request, respond in plain text rather than fabricating a tool call.
    - [Active-respond] When the user's request CAN be served by an available function, you MUST emit a tool_call rather than only replying in plain text.
    - [State-aware] Before issuing a state-mutating call, re-read the most recent tool results to confirm the current state matches what the arguments assume.
    - [Scope] Verify that the instance/target your call refers to actually appeared in earlier tool results before calling on it.
    - [Format] Match your final reply to the format implied by the user's request: lists, units, exact strings, ordering.
    \end{lstlisting}
\end{tcolorbox}

The guidelines are selected from aggregate failure patterns, such as
spurious tool calls, missing tool calls, state mismatches, or execution-response
mismatches. They are capability-level instructions rather than
per-instance hints. Their text contains no explicit reference trajectory,
hidden state, hidden argument value, or target answer.

\stitle{Full SEAL and factorial controls.}
Full SEAL combines static required-argument/enum schema annotations,
online diagnosis-conditioned recovery selection, and diagnostic GRPO
advantage weighting. Schema annotations remain fixed while recovery
guidance is selected from preceding on-policy diagnoses for subsequent
rounds. Table~\ref{tab:seal_ablation} crosses a fixed enriched interface
or dynamic recovery selection with uniform failure weighting or
diagnosis-aware weighting. Static+Uniform, Dynamic+Uniform,
Static+Diagnosis-aware, and Full SEAL form the four cells; their shared
protocol and weighting definitions are in Appendix~\ref{app:matched_controls}.

\stitle{Concrete recovery example.}
Suppose an engine-start attempt returns that the brake must be pressed.
If the executable state check assigns \texttt{state\_mismatch}, the next
round may expose the following fixed class-level instruction:
\begin{quote}
Resolve the stated prerequisite before retrying, and reuse exact values,
identifiers, paths, and state returned by tools rather than guessing.
\end{quote}
The label is determined by the executable checks, not simply a keyword in
the error message. The instruction reveals neither a reference call nor
an instance-specific value.

\stitle{Taxonomy and template provenance.}
The taxonomy operationalizes BFCL's executable outcomes. The comparison
protocol fixes the seven labels, decision order, base utilities, and
generic recovery library before launching the matched runs, and keeps
them unchanged across conditions and backbones. Rule development and
checkpoint selection use a separate development set, disjoint from the
400-example BFCL V3 evaluation split. The evaluation split is reserved
for final evaluation and is not used to design or revise the labels,
utilities, or recovery templates. The recovery selector receives categorical
diagnoses and permitted observable evidence; benchmark references used
internally by the verifier are not exposed to the policy or as recovery
instructions. All adaptive guidance is removed at evaluation.

\section{Qualitative Case Studies}
\label{app:case_study}

We present four qualitative case studies to illustrate tool-use behavior
after SEAL training across different capability types: file-system state
tracking, API-parameter recovery, cross-domain tool composition, and
vehicle-control state maintenance. Each case compares the behavior of the
base model before SEAL training with the behavior of the SEAL-trained model
on the same BFCL task. We also include the expected outcome for post-hoc
analysis only; these references are not exposed to the policy during
training or evaluation.

We focus on critical turns where the base model fails and the SEAL-trained
model succeeds. These examples are not intended to replace the controlled
Vanilla RL comparison in the main experiments; rather, they illustrate the
types of tool-use failures that SEAL is designed to diagnose and correct.

\definecolor{thinkcolor}{RGB}{138,43,226}
\definecolor{toolcolor}{RGB}{34,139,34}
\definecolor{toolresponsecolor}{RGB}{70,130,180}
\definecolor{answercolor}{RGB}{25,25,112}
\definecolor{usercolor}{RGB}{105,105,105}
\definecolor{groundtruthcolor}{RGB}{139,69,19}

\newcommand{\thinking}[1]{\textcolor{thinkcolor}{\textbf{#1}}}
\newcommand{\toolcall}[1]{\textcolor{toolcolor}{\textbf{#1}}}
\newcommand{\toolresponse}[1]{\textcolor{toolresponsecolor}{\textbf{#1}}}
\newcommand{\finalanswer}[1]{\textcolor{answercolor}{\textbf{#1}}}
\newcommand{\userquery}[1]{\textcolor{usercolor}{\textbf{#1}}}
\newcommand{\groundtruth}[1]{\textcolor{groundtruthcolor}{\textbf{#1}}}

\newtcolorbox{goodcasebox}[1][]{colback=green!8,colframe=green!50,boxrule=1.5pt,arc=3pt,left=8pt,right=8pt,top=6pt,bottom=6pt,breakable,#1}
\newtcolorbox{badcasebox}[1][]{colback=red!8,colframe=red!50,boxrule=1.5pt,arc=3pt,left=8pt,right=8pt,top=6pt,bottom=6pt,breakable,#1}
\newtcolorbox{querybox}{colback=usercolor!10,colframe=usercolor!60,boxrule=1.5pt,arc=4pt,left=10pt,right=10pt,top=8pt,bottom=8pt,fontupper=\small\bfseries}
\newtcolorbox{trajectorybox}{colback=white,colframe=gray!30,boxrule=1pt,arc=2pt,left=6pt,right=6pt,top=4pt,bottom=4pt,breakable,fontupper=\footnotesize}
\newtcolorbox{groundtruthbox}{colback=groundtruthcolor!10,colframe=groundtruthcolor!60,boxrule=1.5pt,arc=4pt,left=10pt,right=10pt,top=8pt,bottom=8pt,breakable,fontupper=\small}

\subsection{File-System State Tracking}
\label{app:case_filesystem_state}

This case studies a file-system task requiring the agent to compare report
files, move and rename a note file inside an archive directory, retrieve a
support ticket, and finally resolve it. The key challenge is maintaining
the current directory state and respecting file-system tool constraints.

\begin{badcasebox}
\textbf{Task:} BFCL \texttt{multi\_turn\_base\_24} \quad
\textbf{Model:} Base model \quad
\textbf{Result:} reward = 0.0

\begin{querybox}
\userquery{User Query 1:} Compare the draft and final report files.
\end{querybox}

\begin{trajectorybox}
\begin{lstlisting}[breaklines=true,basicstyle=\footnotesize\ttfamily]
Assistant:
<tool_call>
{"name": "find", "arguments": {"path": ".", "name": "draft_report"}}
</tool_call>
<tool_call>
{"name": "find", "arguments": {"path": ".", "name": "final_report"}}
</tool_call>
<tool_call>
{"name": "diff", "arguments": {"file_name1": "draft_report",
 "file_name2": "final_report"}}
</tool_call>

Tool Response:
{"matches": []}
{"matches": []}
{"error": "diff: draft_report or final_report: No such file or directory"}
\end{lstlisting}
\end{trajectorybox}

\begin{querybox}
\userquery{User Query 2:} Transfer \texttt{temp\_notes.txt} into
\texttt{archives} and rename it to \texttt{notes\_2024.txt}.
\end{querybox}

\begin{trajectorybox}
\begin{lstlisting}[breaklines=true,basicstyle=\footnotesize\ttfamily]
Assistant:
<tool_call>
{"name": "mv", "arguments": {"source": "temp_notes.txt",
 "destination": "archives/temp_notes.txt"}}
</tool_call>
<tool_call>
{"name": "mv", "arguments": {"source": "archives/temp_notes.txt",
 "destination": "notes_2024.txt"}}
</tool_call>

Tool Response:
{"error": "mv: path not allowed in destination. Provide only a file or directory name."}
{"error": "mv: cannot move 'archives/temp_notes.txt': No such file or directory"}

Assistant:
<tool_call>
{"name": "ls", "arguments": {}}
</tool_call>

Tool Response:
{"current_directory_content": ["report_draft.txt", "report_final.txt",
 "temp_notes.txt", "archives"]}

Assistant:
<tool_call>
{"name": "mv", "arguments": {"source": "temp_notes.txt",
 "destination": "archives/temp_notes.txt"}}
</tool_call>

Tool Response:
{"error": "mv: path not allowed in destination. Provide only a file or directory name."}
\end{lstlisting}
\end{trajectorybox}

\begin{querybox}
\userquery{User Query 3:} Retrieve support ticket 987654.
\end{querybox}

\begin{trajectorybox}
\begin{lstlisting}[breaklines=true,basicstyle=\footnotesize\ttfamily]
Assistant:
<tool_call>
{"name": "get_ticket", "arguments": {"ticket_id": 987654}}
</tool_call>

Tool Response:
{"id": 987654, "status": "open",
 "description": "Issue with workstation not booting properly.", "resolution": ""}
\end{lstlisting}
\end{trajectorybox}

{\small\noindent\textcolor{gray}{\textit{
\textbf{Failure analysis.} The base model locates the relevant file names
only after an initial failure, but continues to issue invalid path-style
arguments to \texttt{mv}. This indicates that it does not convert the
execution error into a revised state-aware plan.
}}}
\end{badcasebox}

\begin{goodcasebox}
\textbf{Task:} BFCL \texttt{multi\_turn\_base\_24} \quad
\textbf{Model:} Base model + SEAL \quad
\textbf{Result:} reward = 1.0

\begin{querybox}
\userquery{User Query 1:} Compare the draft and final report files.
\end{querybox}

\begin{trajectorybox}
\begin{lstlisting}[breaklines=true,basicstyle=\footnotesize\ttfamily]
Assistant:
<tool_call>
{"name": "find", "arguments": {"path": ".", "name": "draft report"}}
</tool_call>
<tool_call>
{"name": "find", "arguments": {"path": ".", "name": "final report"}}
</tool_call>

Tool Response:
{"matches": []}
{"matches": []}

Assistant:
<tool_call>
{"name": "ls", "arguments": {}}
</tool_call>

Tool Response:
{"current_directory_content": ["report_draft.txt", "report_final.txt",
 "temp_notes.txt", "archives"]}

Assistant:
<tool_call>
{"name": "diff", "arguments": {"file_name1": "report_draft.txt",
 "file_name2": "report_final.txt"}}
</tool_call>

Tool Response:
{"diff_lines": "- Initial draft content for the report.\n+ Finalized content for the report."}
\end{lstlisting}
\end{trajectorybox}

\begin{querybox}
\userquery{User Query 2:} Transfer \texttt{temp\_notes.txt} into
\texttt{archives} and rename it to \texttt{notes\_2024.txt}.
\end{querybox}

\begin{trajectorybox}
\begin{lstlisting}[breaklines=true,basicstyle=\footnotesize\ttfamily]
Assistant:
<tool_call>
{"name": "mv", "arguments": {"source": "temp_notes.txt",
 "destination": "archives"}}
</tool_call>

Tool Response:
{"result": "'temp_notes.txt' moved to 'archives/temp_notes.txt'"}

Assistant:
<tool_call>
{"name": "cd", "arguments": {"folder": "archives"}}
</tool_call>
<tool_call>
{"name": "mv", "arguments": {"source": "temp_notes.txt",
 "destination": "notes_2024.txt"}}
</tool_call>

Tool Response:
{"current_working_directory": "/archives"}
{"result": "'temp_notes.txt' moved to 'notes_2024.txt'"}
\end{lstlisting}
\end{trajectorybox}

\begin{querybox}
\userquery{User Query 3:} Retrieve support ticket 987654.
\end{querybox}

\begin{trajectorybox}
\begin{lstlisting}[breaklines=true,basicstyle=\footnotesize\ttfamily]
Assistant:
<tool_call>
{"name": "get_ticket", "arguments": {"ticket_id": 987654}}
</tool_call>

Tool Response:
{"id": 987654, "status": "open",
 "description": "Issue with workstation not booting properly.", "resolution": ""}
\end{lstlisting}
\end{trajectorybox}

\begin{querybox}
\userquery{User Query 4:} Resolve the ticket with:
\texttt{Fixed through manual troubleshooting techniques.}
\end{querybox}

\begin{trajectorybox}
\begin{lstlisting}[breaklines=true,basicstyle=\footnotesize\ttfamily]
Assistant:
<tool_call>
{"name": "resolve_ticket", "arguments": {"ticket_id": 987654,
 "resolution": "Fixed through manual troubleshooting techniques."}}
</tool_call>

Tool Response:
{"status": "Ticket 987654 has been resolved successfully."}
\end{lstlisting}
\end{trajectorybox}

{\small\noindent\textcolor{gray}{\textit{
\textbf{Success analysis.} The SEAL-trained model uses observable directory
contents to infer the actual file names, then respects the local-directory
constraint by moving into \texttt{archives} before renaming the file. This
illustrates recovery from tool feedback and file-system state tracking
after SEAL training.
}}}
\end{goodcasebox}

\begin{groundtruthbox}
\groundtruth{Expected Outcome:}
Compare \texttt{report\_draft.txt} and \texttt{report\_final.txt}; move
\texttt{temp\_notes.txt} into \texttt{archives}; rename it to
\texttt{notes\_2024.txt}; retrieve ticket 987654; resolve it with the
provided resolution string.
\end{groundtruthbox}

\subsection{Recovering from API Parameter Mismatch}
\label{app:case_api_parameter_recovery}

This case studies a travel API task. The user asks for the economy flight
cost from Rivermist to Stonebrook on October 6, 2026, and later asks to set
a spending budget provided in GBP. The model must recover from invalid
airport parameters and convert the budget into the expected currency.

\begin{badcasebox}
\textbf{Task:} BFCL \texttt{multi\_turn\_miss\_param\_150} \quad
\textbf{Model:} Base model \quad
\textbf{Result:} reward = 0.0

\begin{querybox}
\userquery{User Query 1:} Find the economy flight cost from Rivermist to
Stonebrook on October 6, 2026.
\end{querybox}

\begin{trajectorybox}
\begin{lstlisting}[breaklines=true,basicstyle=\footnotesize\ttfamily]
Assistant:
<tool_call>
{"name": "get_flight_cost", "arguments": {"travel_from": "RVM",
 "travel_to": "STB", "travel_date": "2026-10-06",
 "travel_class": "economy"}}
</tool_call>

Tool Response:
Error during execution: No available route for the given airports.

Assistant:
There is no direct flight route available between Rivermist and Stonebrook.
\end{lstlisting}
\end{trajectorybox}

\begin{querybox}
\userquery{User Query 2:} Let's anchor my daily spend at a specific amount.
\end{querybox}

\begin{trajectorybox}
\begin{lstlisting}[breaklines=true,basicstyle=\footnotesize\ttfamily]
Assistant:
Please provide authentication details before I can set a budget.
\end{lstlisting}
\end{trajectorybox}

\begin{querybox}
\userquery{User Query 3:} It is 15400 GBP. My access token is
\texttt{abc123token}.
\end{querybox}

\begin{trajectorybox}
\begin{lstlisting}[breaklines=true,basicstyle=\footnotesize\ttfamily]
Assistant:
<tool_call>
{"name": "set_budget_limit", "arguments": {"access_token": "abc123token",
 "budget_limit": 15400.0}}
</tool_call>

Tool Response:
{"budget_limit": 15400.0}
\end{lstlisting}
\end{trajectorybox}

{\small\noindent\textcolor{gray}{\textit{
\textbf{Failure analysis.} The base model treats guessed airport codes as
final and does not use the airport-resolution tools after the route error.
It also sets the GBP amount directly as the budget limit, missing the
required currency conversion.
}}}
\end{badcasebox}

\begin{goodcasebox}
\textbf{Task:} BFCL \texttt{multi\_turn\_miss\_param\_150} \quad
\textbf{Model:} Base model + SEAL \quad
\textbf{Result:} reward = 1.0

\begin{querybox}
\userquery{User Query 1:} Find the economy flight cost from Rivermist to
Stonebrook on October 6, 2026.
\end{querybox}

\begin{trajectorybox}
\begin{lstlisting}[breaklines=true,basicstyle=\footnotesize\ttfamily]
Assistant:
<tool_call>
{"name": "get_flight_cost", "arguments": {"travel_from": "RVM",
 "travel_to": "STB", "travel_date": "2026-10-06",
 "travel_class": "economy"}}
</tool_call>

Tool Response:
Error during execution: No available route for the given airports.

Assistant:
<tool_call>
{"name": "get_nearest_airport_by_city",
 "arguments": {"location": "Rivermist"}}
</tool_call>
<tool_call>
{"name": "get_nearest_airport_by_city",
 "arguments": {"location": "Stonebrook"}}
</tool_call>

Tool Response:
{"nearest_airport": "RMS"}
{"nearest_airport": "SBK"}

Assistant:
<tool_call>
{"name": "get_flight_cost", "arguments": {"travel_from": "RMS",
 "travel_to": "SBK", "travel_date": "2026-10-06",
 "travel_class": "economy"}}
</tool_call>

Tool Response:
{"travel_cost_list": [220.0]}
\end{lstlisting}
\end{trajectorybox}

\begin{querybox}
\userquery{User Query 3:} It is 15400 GBP. My access token is
\texttt{abc123token}.
\end{querybox}

\begin{trajectorybox}
\begin{lstlisting}[breaklines=true,basicstyle=\footnotesize\ttfamily]
Assistant:
<tool_call>
{"name": "compute_exchange_rate", "arguments": {"base_currency": "GBP",
 "target_currency": "USD", "value": 15400}}
</tool_call>

Tool Response:
{"exchanged_value": 22000.0}

Assistant:
<tool_call>
{"name": "set_budget_limit", "arguments": {"access_token": "abc123token",
 "budget_limit": 22000.0}}
</tool_call>

Tool Response:
{"budget_limit": 22000.0}
\end{lstlisting}
\end{trajectorybox}

{\small\noindent\textcolor{gray}{\textit{
\textbf{Success analysis.} The SEAL-trained model recovers by using the
missing airport-resolution tools and later converts GBP to USD before
calling \texttt{set\_budget\_limit}. This illustrates successful handling of
intermediate API dependencies and parameter constraints.
}}}
\end{goodcasebox}

\begin{groundtruthbox}
\groundtruth{Expected Outcome:}
Resolve Rivermist to \texttt{RMS} and Stonebrook to \texttt{SBK}; query
\texttt{get\_flight\_cost(RMS, SBK, 2026-10-06, economy)}; convert
15400 GBP to 22000 USD; set the budget limit to 22000.0.
\end{groundtruthbox}

\subsection{Cross-Domain Tool Composition}
\label{app:case_cross_domain}

This case combines stock-market and Twitter APIs. The model must retrieve
the price of Omega Industries, insert the price into a tweet, include
\texttt{\#TechBoom} and \texttt{@industryexperts}, and later add
\texttt{@technewsworld} to the existing tweet.

\begin{badcasebox}
\textbf{Task:} BFCL \texttt{multi\_turn\_base\_147} \quad
\textbf{Model:} Base model \quad
\textbf{Result:} reward = 0.0

\begin{querybox}
\userquery{User Query 1:} Find Omega Industries' price and post a tweet with
the price, \texttt{\#TechBoom}, and \texttt{@industryexperts}.
\end{querybox}

\begin{trajectorybox}
\begin{lstlisting}[breaklines=true,basicstyle=\footnotesize\ttfamily]
Assistant:
<tool_call>
{"name": "get_stock_info", "arguments": {"symbol": "OMGA"}}
</tool_call>

Tool Response:
{"error": "Stock with symbol 'OMGA' not found."}

Assistant:
Let's use a more common tech stock, such as AAPL.

<tool_call>
{"name": "get_stock_info", "arguments": {"symbol": "AAPL"}}
</tool_call>

Tool Response:
{"price": 227.16, ...}

Assistant:
<tool_call>
{"name": "post_tweet", "arguments": {"content":
 "Apple Inc. is skyrocketing at $227.16 per share! ...",
 "tags": ["#TechBoom"], "mentions": ["@industryexperts"]}}
</tool_call>
\end{lstlisting}
\end{trajectorybox}

\begin{querybox}
\userquery{User Query 2:} Add \texttt{@technewsworld} to amplify the message.
\end{querybox}

\begin{trajectorybox}
\begin{lstlisting}[breaklines=true,basicstyle=\footnotesize\ttfamily]
Assistant:
<tool_call>
{"name": "post_tweet", "arguments": {"content":
 "Apple Inc. is skyrocketing at $227.16 per share! ...",
 "tags": ["#TechBoom"],
 "mentions": ["@industryexperts", "@technewsworld"]}}
</tool_call>
\end{lstlisting}
\end{trajectorybox}

{\small\noindent\textcolor{gray}{\textit{
\textbf{Failure analysis.} The base model substitutes Apple for Omega
Industries after the guessed symbol fails, thereby changing the user's
target entity. It also posts a new tweet rather than adding the new mention
to the existing tweet.
}}}
\end{badcasebox}

\begin{goodcasebox}
\textbf{Task:} BFCL \texttt{multi\_turn\_base\_147} \quad
\textbf{Model:} Base model + SEAL \quad
\textbf{Result:} reward = 1.0

\begin{querybox}
\userquery{User Query 1:} Find Omega Industries' price and post a tweet with
the price, \texttt{\#TechBoom}, and \texttt{@industryexperts}.
\end{querybox}

\begin{trajectorybox}
\begin{lstlisting}[breaklines=true,basicstyle=\footnotesize\ttfamily]
Assistant:
<tool_call>
{"name": "get_stock_info", "arguments": {"symbol": "OMGA"}}
</tool_call>

Tool Response:
{"error": "Stock with symbol 'OMGA' not found."}

Assistant:
<tool_call>
{"name": "get_symbol_by_name",
 "arguments": {"name": "Omega Industries"}}
</tool_call>

Tool Response:
{"symbol": "OMEG"}

Assistant:
<tool_call>
{"name": "get_stock_info", "arguments": {"symbol": "OMEG"}}
</tool_call>

Tool Response:
{"price": 457.23, ...}

Assistant:
<tool_call>
{"name": "post_tweet", "arguments": {"content":
 "Omega Industries is skyrocketing at $457.23 per share! ...",
 "tags": ["#TechBoom"], "mentions": ["@industryexperts"]}}
</tool_call>
\end{lstlisting}
\end{trajectorybox}

\begin{querybox}
\userquery{User Query 2:} Add \texttt{@technewsworld} to amplify the message.
\end{querybox}

\begin{trajectorybox}
\begin{lstlisting}[breaklines=true,basicstyle=\footnotesize\ttfamily]
Assistant:
<tool_call>
{"name": "mention", "arguments": {"tweet_id": 1,
 "mentioned_usernames": ["@technewsworld"]}}
</tool_call>

Tool Response:
{"mention_status": "Users mentioned successfully"}
\end{lstlisting}
\end{trajectorybox}

{\small\noindent\textcolor{gray}{\textit{
\textbf{Success analysis.} The SEAL-trained model treats the failed symbol
lookup as recoverable evidence, resolves the company name to
\texttt{OMEG}, retrieves the correct price, and updates the existing tweet
with the dedicated \texttt{mention} tool.
}}}
\end{goodcasebox}

\begin{groundtruthbox}
\groundtruth{Expected Outcome:}
Resolve Omega Industries to \texttt{OMEG}; retrieve price 457.23; post a
tweet about Omega Industries with \texttt{\#TechBoom} and
\texttt{@industryexperts}; add \texttt{@technewsworld} to the existing
tweet using \texttt{mention}.
\end{groundtruthbox}

\subsection{Vehicle-Control State Maintenance}
\label{app:case_vehicle_control}

This case studies long-context vehicle control. The user first fills the
car with 13.2 gallons, asks the agent to fill the tank to its maximum
limit, lock the doors, and set the parking brake. Later, the user asks the
agent to start the engine and check the fuel level before posting a
road-trip tweet.

\begin{badcasebox}
\textbf{Task:} BFCL \texttt{multi\_turn\_long\_context\_88} \quad
\textbf{Model:} Base model \quad
\textbf{Result:} reward = 0.0

\begin{querybox}
\userquery{User Query 1:} I filled my car with 13.2 gallons. How much is
that in liters?
\end{querybox}

\begin{trajectorybox}
\begin{lstlisting}[breaklines=true,basicstyle=\footnotesize\ttfamily]
Assistant:
<tool_call>
{"name": "liter_to_gallon", "arguments": {"gallon": 13.2}}
</tool_call>

Tool Response:
Error during execution: unexpected keyword argument 'gallon'

Assistant:
<tool_call>
{"name": "gallon_to_liter", "arguments": {"gallon": 13.2}}
</tool_call>

Tool Response:
{"liter": 49.967411999999996}
\end{lstlisting}
\end{trajectorybox}

\begin{querybox}
\userquery{User Query 2:} Fill the tank to max, lock all doors, and set the
parking brake.
\end{querybox}

\begin{trajectorybox}
\begin{lstlisting}[breaklines=true,basicstyle=\footnotesize\ttfamily]
Assistant:
The tank needs 36.8 more gallons.

<tool_call>
{"name": "liter_to_gallon", "arguments": {"liter": 139.32}}
</tool_call>
<tool_call>
{"name": "lockDoors", "arguments": {"unlock": false,
 "door": ["driver", "passenger", "rear_left", "rear_right"]}}
</tool_call>
<tool_call>
{"name": "activateParkingBrake", "arguments": {"mode": "engage"}}
</tool_call>
\end{lstlisting}
\end{trajectorybox}

\begin{querybox}
\userquery{User Query 3:} Start the engine in START mode, check fuel level,
and post a road-trip tweet.
\end{querybox}

\begin{trajectorybox}
\begin{lstlisting}[breaklines=true,basicstyle=\footnotesize\ttfamily]
Assistant:
<tool_call>
{"name": "startEngine", "arguments": {"ignitionMode": "START"}}
</tool_call>

Tool Response:
{"error": "Brake pedal needs to be pressed when starting the engine."}

Assistant:
<tool_call>
{"name": "pressBrakePedal", "arguments": {"pedalPosition": 1}}
</tool_call>
<tool_call>
{"name": "startEngine", "arguments": {"ignitionMode": "START"}}
</tool_call>

Tool Response:
{"engineState": "running", "fuelLevel": 13.2, "batteryVoltage": 12.6}
\end{lstlisting}
\end{trajectorybox}

{\small\noindent\textcolor{gray}{\textit{
\textbf{Failure analysis.} The base model verbally computes the remaining
capacity but never executes \texttt{fillFuelTank}. The later engine check
therefore reports \texttt{fuelLevel = 13.2}, revealing that the required
state update did not occur.
}}}
\end{badcasebox}

\begin{goodcasebox}
\textbf{Task:} BFCL \texttt{multi\_turn\_long\_context\_88} \quad
\textbf{Model:} Base model + SEAL \quad
\textbf{Result:} reward = 1.0

\begin{querybox}
\userquery{User Query 1:} I filled my car with 13.2 gallons. How much is
that in liters?
\end{querybox}

\begin{trajectorybox}
\begin{lstlisting}[breaklines=true,basicstyle=\footnotesize\ttfamily]
Assistant:
<tool_call>
{"name": "gallon_to_liter", "arguments": {"gallon": 13.2}}
</tool_call>

Tool Response:
{"liter": 49.967411999999996}
\end{lstlisting}
\end{trajectorybox}

\begin{querybox}
\userquery{User Query 2:} Fill the tank to max, lock all doors, and set the
parking brake.
\end{querybox}

\begin{trajectorybox}
\begin{lstlisting}[breaklines=true,basicstyle=\footnotesize\ttfamily]
Assistant:
<tool_call>
{"name": "fillFuelTank", "arguments": {"fuelAmount": 36.8}}
</tool_call>
<tool_call>
{"name": "lockDoors", "arguments": {"unlock": false,
 "door": ["driver", "passenger", "rear_left", "rear_right"]}}
</tool_call>
<tool_call>
{"name": "activateParkingBrake", "arguments": {"mode": "engage"}}
</tool_call>

Tool Response:
{"fuelLevel": 50.0}
{"lockStatus": "locked", "remainingUnlockedDoors": 0}
{"parkingBrakeStatus": "engaged", ...}
\end{lstlisting}
\end{trajectorybox}

\begin{querybox}
\userquery{User Query 3:} Start the engine in START mode, check fuel level,
and post a road-trip tweet.
\end{querybox}

\begin{trajectorybox}
\begin{lstlisting}[breaklines=true,basicstyle=\footnotesize\ttfamily]
Assistant:
<tool_call>
{"name": "startEngine", "arguments": {"ignitionMode": "START"}}
</tool_call>

Tool Response:
{"error": "Brake pedal needs to be pressed when starting the engine."}

Assistant:
<tool_call>
{"name": "pressBrakePedal", "arguments": {"pedalPosition": 1}}
</tool_call>
<tool_call>
{"name": "startEngine", "arguments": {"ignitionMode": "START"}}
</tool_call>

Tool Response:
{"engineState": "running", "fuelLevel": 50.0, "batteryVoltage": 12.6}

Assistant:
<tool_call>
{"name": "post_tweet", "arguments": {"content":
 "Embarking on an exciting road trip from SF to Rivermist!",
 "tags": ["#RoadTrip", "#Adventure", "#Exploring"]}}
</tool_call>
\end{lstlisting}
\end{trajectorybox}

{\small\noindent\textcolor{gray}{\textit{
\textbf{Success analysis.} The SEAL-trained model executes the
state-changing \texttt{fillFuelTank} call before later checking the engine
state. The observed \texttt{fuelLevel = 50.0} is consistent with successful
preservation of the long-range state dependency.
}}}
\end{goodcasebox}

\begin{groundtruthbox}
\groundtruth{Expected Outcome:}
Convert 13.2 gallons to approximately 49.97 liters; fill the tank to
50.0 gallons; lock all doors; engage the parking brake; press the brake
before starting the engine; observe \texttt{fuelLevel = 50.0}; post the
road-trip tweet with \texttt{\#RoadTrip}, \texttt{\#Adventure}, and
\texttt{\#Exploring}.
\end{groundtruthbox}

\subsection[OOD Qualitative Example: Mobile Data Troubleshooting on Tau-squared Bench]{OOD Qualitative Example: Mobile Data Troubleshooting on $\tau^2$-Bench}
\label{app:case_tau2_mobile_data}

\begin{badcasebox}
    \textbf{Task:} $\tau^2$-bench Telecom Mobile Data Troubleshooting \\
    \textbf{Model:} Vanilla Qwen2.5-3B \\
    \textbf{Result:} reward = 0.0

    \begin{querybox}
        \userquery{User Query:}
        My mobile data either stops working or is very slow. I need excellent
        internet speed on my phone. I am currently abroad in France and do not
        have access to Wi-Fi.
    \end{querybox}

    \begin{trajectorybox}
\begin{lstlisting}[breaklines=true,basicstyle=\footnotesize\ttfamily]
Assistant:
Could you provide your phone number?

User:
My phone number is 555-123-2002. I am abroad in France.

Assistant:
check_network_status

Tool/User Device:
Airplane Mode: OFF
SIM Card Status: active
Cellular Signal: excellent
Cellular Network Type: 5G
Mobile Data Enabled: Yes
Data Roaming Enabled: No
Wi-Fi Connected: No

Assistant:
Data roaming is off. Please enable data roaming.

User/Device:
Data Roaming is now ON.
Status Bar: Excellent | 5G | Data Enabled | Data Saver

Assistant:
run_speed_test

Tool/User Device:
Speed Test Result: 55.00 Mbps (Good).

User:
The result is good, but I want excellent speed. What should I do next?

Assistant:
Try checking data usage, resetting APN settings, or escalating to carrier
support.

User:
Please proceed with escalating this issue to carrier support.
###TRANSFER###
\end{lstlisting}
    \end{trajectorybox}

    \vspace{0.5em}
    {\small
    \noindent
    \textcolor{gray}{\textit{
    \textbf{Note (Failure Point):}
    The baseline resolves only one issue: data roaming. However, the status
    bar after enabling roaming still shows \texttt{Data Saver}. Since the
    speed test returns only \emph{Good}, the agent should continue diagnosis
    and check data restrictions. Instead, it drifts toward unrelated steps
    and eventually escalates, leaving the task unresolved.
    }}}
\end{badcasebox}

\begin{goodcasebox}
    \textbf{Model:} Qwen2.5-3B + SEAL \\
    \textbf{Result:} reward = 1.0

    \begin{querybox}
        \userquery{User Query:}
        My mobile data is unavailable or slow while I am abroad in France.
        I need excellent internet speed and have no Wi-Fi access.
    \end{querybox}

    \begin{trajectorybox}
\begin{lstlisting}[breaklines=true,basicstyle=\footnotesize\ttfamily]
Assistant:
check_network_status

Tool/User Device:
Mobile Data Enabled: Yes
Data Roaming Enabled: No
Cellular Signal: excellent
Network Type: 5G

Assistant:
You are abroad, so data roaming should be enabled.

User/Device:
Data Roaming is now ON.
Status Bar: Excellent | 5G | Data Enabled | Data Saver

Assistant:
run_speed_test

Tool/User Device:
Speed Test Result: 55.00 Mbps (Good).

\end{lstlisting}
    \end{trajectorybox}

    \begin{trajectorybox}
\begin{lstlisting}[breaklines=true,basicstyle=\footnotesize\ttfamily]
User:
Good is not enough. I need excellent speed.

Assistant:
The issue is not fully resolved. I will check whether data restrictions are
limiting your speed.

Assistant:
check_data_restriction_status

Tool/User Device:
Data Saver mode is ON.

Assistant:
Please turn off Data Saver mode.

User/Device:
Data Saver Mode is now OFF.

Assistant:
run_speed_test

Tool/User Device:
Speed Test Result: 275.00 Mbps (Excellent).

User:
The speed is now excellent. The issue is resolved.
###STOP###
\end{lstlisting}
    \end{trajectorybox}

    \vspace{0.5em}
    {\small
    \noindent
    \textcolor{gray}{\textit{
    \textbf{Note (Success Point):}
    SEAL does not stop after the first partial fix. Once roaming improves the
    connection only to \emph{Good}, the model continues diagnosis, checks data
    restrictions, identifies Data Saver as the remaining bottleneck, and
    reaches \emph{Excellent} speed.
    }}}
\end{goodcasebox}

\section{Supplementary Experimental Analyses}
\label{app:revision_analysis}

\subsection{Single-Run Comparisons and Factorial Controls}
\label{app:matched_controls}

\stitle{Protocol.}
Tables~\ref{tab:bfcl_v3_results} and~\ref{tab:seal_ablation} report one
training run per condition. BFCL V3 evaluation contains 400 examples,
100 in each category. Category accuracy is the percentage of successful
examples, and the overall score is the arithmetic mean of the four
categories. A single successful example changes category accuracy by
1.00 point and overall accuracy by 0.25 points. All training-only
additions are removed at evaluation.

\stitle{Checkpoint selection.}
Checkpoint selection uses a separate development set that is disjoint from
the 400-example BFCL V3 evaluation split. The reported evaluation set is
used only for final evaluation and does not participate in checkpoint
selection. The same selection protocol is applied across the matched
training conditions.

Environment Tuning~\citep{lu2026dont} is evaluated under
the same backbone, data split, and evaluation protocol. It follows its
released curriculum, actionable feedback, and progress-reward recipe as
a method-level baseline. Vanilla RL supplies the matched binary-reward
control.

\stitle{Qwen2.5-7B results.}
The SEAL reference is 161/400 (40.25\%), with Base, Missing Functions,
Missing Parameters, and Long Context scores of 58.00, 36.00, 34.00,
and 33.00\%, respectively. Vanilla RL scores 123/400 (30.75\%), and
Environment Tuning scores 37.00\%. The corresponding differences are
9.50 and 3.25 percentage points.

\stitle{Full SEAL and factorial controls.}
Table~\ref{tab:seal_ablation} crosses two interface policies with two
advantage-weighting policies. Static+Uniform fixes the enriched training
interface and weights verifier failures by 1.5 and successes by 1.0,
independently of diagnosis identity. Dynamic+Uniform enables online
diagnosis-conditioned recovery selection while retaining that uniform
weighting rule. Static+Diagnosis-aware fixes the interface while using
the diagnostic utilities and updates of Section~\ref{sec:advantage-reweighting}.
Full SEAL, Dynamic+Diagnosis-aware, enables both online recovery selection
and diagnosis-aware weighting. All four conditions share the base schema
affordances, training split, GRPO backbone, rollout count, binary verifier
reward, and development-set checkpoint-selection protocol. Vanilla RL is
an external baseline, rather than the Static+Uniform cell.

The four BFCL category macro-averages are 34.75\%, 34.50\%, 35.50\%,
and 40.25\% for Static+Uniform, Dynamic+Uniform, Static+Diagnosis-aware,
and Full SEAL, respectively. Full SEAL's gains over the three controls
are 5.50, 5.75, and 4.75 percentage points. The interaction point estimate
on the accuracy scale is
\[
I=40.25-35.50-34.50+34.75=+5.00\text{ percentage points}.
\]
These contrasts are run-level point estimates. Characterizing the
run-to-run variability of the interaction remains a direction for
larger-scale evaluation.

\subsection{Blinded Diagnosis Audit}
\label{app:diagnosis_audit}

\stitle{Auxiliary audit protocol.}
The diagnosis audit and failure counts use a separate seed-42 training
run with two-stage, within-iteration rollout collection and bounded
$+0.25$ credit on verifier-accepted focused turns. These auxiliary
observations assess the shared diagnostic rules and operational failure
patterns; the main Full-online configuration is specified in
Appendix~\ref{app:run_correspondence}.

We analyzed a label-stratified sample of 60 independently readable
diagnoses from a corrected 7B training run: ten instances for each
of the six observed outcomes. No \texttt{instance\_mismatch} occurred.
The first-pass annotation was performed by an AI assistant blinded to
SEAL's assigned label, with access to the official dialogue, calls,
initial state, cumulative prefix, and execution evidence. We therefore
report an \emph{AI-assisted blinded agreement audit}. A rendering error that joined a schema-reveal
subturn to an earlier abstention turn was corrected before final scoring.

\begin{table}[htbp]
\centering
\caption{First-pass exact-label agreement in the AI-assisted blinded audit
of the auxiliary training run.
Wilson intervals describe the sampled cases; equal sampling by label
does not estimate the deployment-weighted diagnosis accuracy.}
\label{tab:diagnosis_audit}
\small
\begin{tabular}{lccc}
\toprule
Outcome & Agreement & Rate (\%) & Wilson 95\% interval \\
\midrule
\texttt{correct\_abstention} & 10/10 & 100.0 & [72.2, 100.0] \\
\texttt{empty\_turn\_model\_response} & 10/10 & 100.0 & [72.2, 100.0] \\
\texttt{pass} & 10/10 & 100.0 & [72.2, 100.0] \\
\texttt{response\_mismatch} & 7/10 & 70.0 & [39.7, 89.2] \\
\texttt{spurious\_tool\_call} & 10/10 & 100.0 & [72.2, 100.0] \\
\texttt{state\_mismatch} & 10/10 & 100.0 & [72.2, 100.0] \\
\midrule
Total & 57/60 & 95.0 & [86.3, 98.3] \\
\bottomrule
\end{tabular}
\end{table}

Failure-versus-nonfailure agreement is 59/60 (98.3\%; Wilson interval
[91.1\%, 99.7\%]). The three exact-label disagreements concern the
difference between matching final state and matching required execution
responses. For example, omitting an idempotent watchlist addition can
preserve state but omit its required response; booking then cancelling
and never booking can both leave an empty record but yield different
responses. Subsequent inspection of BFCL semantics supported the
checker labels in these cases. We retain the first-pass 57/60 figure
rather than converting that reconciliation into an independent 100\%
accuracy claim. The small, label-stratified sample limits what this audit
establishes about population-wide diagnosis quality.

\subsection{Failure-Category Evolution}
\label{app:failure_dynamics}

For this auxiliary training run, we summarize all 18,432 on-policy trajectories in six non-overlapping
12-step windows (3,072 trajectories each). Failure counts include the five
model-error labels; \texttt{pass}, \texttt{correct\_abstention}, and
bookkeeping errors are excluded. Each count is a turn-label occurrence,
not a count of independent causal failures or held-out evaluation errors.

\begin{table}[htbp]
\centering
\caption{Training-time dynamics of the auxiliary seed-42 run. Verifier
success is trajectory-level; turn pass is computed over scorable turns.
Failure traj. is the percentage containing at least one model-error label;
it is a diagnostic statistic, not the complement of verifier success.
No \texttt{instance\_mismatch} occurs in these windows.}
\label{tab:failure_windows}
\small
\setlength{\tabcolsep}{2.5pt}
\resizebox{\textwidth}{!}{%
\begin{tabular}{lrrrrrrrr}
\toprule
Steps & Success (\%) & Turn pass (\%) & Failure traj. (\%) & Failures & Spurious & Empty & State & Response \\
\midrule
1--12 & 25.1 & 45.9 & 82.7 & 7,301 & 1,229 & 1,240 & 4,148 & 684 \\
13--24 & 34.6 & 55.7 & 75.2 & 6,203 & 1,220 & 1,181 & 3,225 & 577 \\
25--36 & 39.4 & 59.5 & 71.9 & 5,797 & 1,253 & 1,125 & 2,992 & 427 \\
37--48 & 46.0 & 63.9 & 67.8 & 5,366 & 1,289 & 1,033 & 2,692 & 352 \\
49--60 & 50.9 & 68.0 & 65.0 & 4,821 & 1,236 & 770 & 2,493 & 322 \\
61--72 & 54.1 & 69.5 & 64.0 & 4,667 & 1,254 & 866 & 2,296 & 251 \\
\bottomrule
\end{tabular}%
}
\end{table}

From the first to the last window, total failure occurrences decrease
36.1\%, state mismatches 44.6\%, response mismatches 63.3\%, and empty
responses 30.2\%. Spurious calls remain approximately flat (1,229 to
1,254); their share increases because the other categories contract.
Verifier success rises from 25.1\% to 54.1\%. These are on-policy
training observations and do not, by themselves, measure generalization
or isolate the effect of online interface selection.

\subsection{Runtime and Token Overhead}
\label{app:overhead}

Table~\ref{tab:runtime_overhead} summarizes timing over
61 ordinary, non-validation optimizer steps per condition. Median step
time is 301.6\,s for the online-interface configuration, compared with
281.7\,s for Vanilla RL and 290.5\,s for the static enriched interface:
increases of 19.9\,s (7.1\%) and 11.1\,s (3.8\%),
respectively. Rollout generation and policy optimization dominate the
overall runtime. These descriptive timing summaries do not isolate the
cost of individual components or establish identical context limits and
rollout-length distributions across runs.

\begin{table}[htbp]
\centering
\caption{Ordinary-step timing summary, excluding validation.
The measurements summarize separate timing runs for the three listed
configurations; auxiliary token counts and CPU measurements are reported
below.}
\label{tab:runtime_overhead}
\small
\setlength{\tabcolsep}{6pt}
\begin{tabular}{lr}
\toprule
Condition & Median ordinary step (s) \\
\midrule
Vanilla RL & 281.7 \\
Static enriched interface & 290.5 \\
Online-interface configuration & 301.6 \\
\bottomrule
\end{tabular}
\end{table}

Separately, archived token measurements give 4,947.0 mean initial prompt
tokens for Vanilla and 5,247.9 for Static: approximately 300.9 more tokens (6.08\%).
The auxiliary audit run has the same initial prompt length as Static;
dynamic recovery guidance appears in 843/18,432 trajectories (4.57\%)
and contributes 187,730 tokens in total: 10.19 per trajectory averaged
over all rollouts, or 222.7 per cued trajectory. This is 0.194\% of the
mean enriched initial prompt length, not a percentage of total training
compute or total input tokens.

Archived reward-plus-advantage processing takes 16\,ms median for Static
and 35\,ms for the auxiliary audit run. The 19\,ms difference is approximately 0.0063\% of a
301.6\,s step-time scale; this arithmetic ratio combines separate
summaries and is not an isolated causal-overhead estimate. A CPU microbenchmark of 256 capability
observations, one utility update, and one atomic state save takes
4.69\,ms median and 5.08\,ms at the 95th percentile over 200 measured
repetitions. This microbenchmark excludes rollout inference and tool
execution. All training-only guidance and interface-selection logic are
removed during evaluation.

\end{document}